%% file: main.tex
\documentclass[nohyperref]{article}

\usepackage{microtype}
\usepackage{graphicx}
\usepackage{subfigure}
\usepackage{booktabs} % for professional tables

\usepackage{hyperref}
\usepackage{adjustbox}
\usepackage{url}
\usepackage{multirow}
\usepackage[normalem]{ulem}
\useunder{\uline}{\ul}{}
\usepackage[ruled,vlined, linesnumbered]{algorithm2e}
\usepackage[numbers]{natbib}

\newcommand{\highest}[1]{\textbf{\textcolor{blue}{#1}}}

\usepackage[final]{src/nips2022}

\usepackage{amsmath}
\usepackage{amssymb}
\usepackage{mathtools}
\usepackage{amsthm}
\usepackage{tabularx}
\usepackage{pifont}

\usepackage[capitalize,noabbrev]{cleveref}

\theoremstyle{plain}

\theoremstyle{definition}

\theoremstyle{remark}

\usepackage[utf8]{inputenc} % allow utf-8 input
\usepackage[T1]{fontenc}    % use 8-bit T1 fonts
\usepackage{hyperref}       % hyperlinks
\usepackage{url}            % simple URL typesetting
\usepackage{booktabs}       % professional-quality tables
\usepackage{amsfonts}       % blackboard math symbols
\usepackage{nicefrac}       % compact symbols for 1/2, etc.
\usepackage{microtype}      % microtypography
\usepackage{xcolor}         % colors

\title{Paraphrasing Is All You Need\\
for Novel Object Captioning}

\usepackage[textsize=tiny]{todonotes}

\author{%
  Cheng-Fu Yang$^{1}$ \qquad Yao-Hung Hubert Tsai$^{2}$ \qquad Wan-Cyuan Fan$^{3}$ \AND Ruslan Salakhutdinov$^{2}$ \qquad Louis-Philippe Morency$^{2}$ \qquad Yu-Chiang Frank Wang$^{3,4}$ \\
  \\
  $^{1}$UCLA \qquad $^{2}$Carnegie Mellon University \qquad
  $^{3}$National Taiwan University \qquad
  $^{4}$NVIDIA \\
  \\
  \texttt{cfyang@cs.ucla.edu} \\
}

\begin{document}
\maketitle

\begin{abstract}
\input{sections/0_abstract}
\end{abstract}

\input{sections/1_intro}

\input{sections/5_related}
% \input{sections/2_connection}
\input{sections/3_method}
\input{sections/4_exp}
\input{sections/6_conclusion}

\bibliography{iclr2022_conference}
\bibliographystyle{unsrtnat}

\section*{Checklist}

\begin{enumerate}

\item For all authors...
\begin{enumerate}
  \item Do the main claims made in the abstract and introduction accurately reflect the paper's contributions and scope?
    \answerYes{}
  \item Did you describe the limitations of your work?
    \answerYes{In our supplementary materials, we demonstrate some failure cases of our model. We empirically observe that the error mainly comes from the wrong detection tags predicted by the pre-trained object detectors.}
  \item Did you discuss any potential negative societal impacts of your work?
    \answerNo{}
  \item Have you read the ethics review guidelines and ensured that your paper conforms to them?
    \answerYes{}
\end{enumerate}

\item If you are including theoretical results...
\begin{enumerate}
  \item Did you state the full set of assumptions of all theoretical results?
    \answerNA{}
        \item Did you include complete proofs of all theoretical results?
    \answerNA{}
\end{enumerate}

\item If you ran experiments...final
\begin{enumerate}
  \item Did you include the code, data, and instructions needed to reproduce the main experimental results (either in the supplemental material or as a URL)?
    \answerYes{}
  \item Did you specify all the training details (e.g., data splits, hyperparameters, how they were chosen)?
    \answerYes{Please see the supplementary materials.}
        \item Did you report error bars (e.g., with respect to the random seed after running experiments multiple times)?
    \answerNo{}
        \item Did you include the total amount of compute and the type of resources used (e.g., type of GPUs, internal cluster, or cloud provider)?
    \answerYes{}
\end{enumerate}

\item If you are using existing assets (e.g., code, data, models) or curating/releasing new assets...
\begin{enumerate}
  \item If your work uses existing assets, did you cite the creators?
    \answerYes{}
  \item Did you mention the license of the assets?
    \answerYes{}
  \item Did you include any new assets either in the supplemental material or as a URL?
    \answerNo{}
  \item Did you discuss whether and how consent was obtained from people whose data you're using/curating?
    \answerYes{}{See Section~\ref{sec:exp}. All datasets we used are public.}
  \item Did you discuss whether the data you are using/curating contains personally identifiable information or offensive content?
    \answerNA{}
\end{enumerate}

\item If you used crowdsourcing or conducted research with human subjects...
\begin{enumerate}
  \item Did you include the full text of instructions given to participants and screenshots, if applicable?
    \answerYes{Please see our supplementary materials.}
  \item Did you describe any potential participant risks, with links to Institutional Review Board (IRB) approvals, if applicable?
    \answerNo{}
  \item Did you include the estimated hourly wage paid to participants and the total amount spent on participant compensation?
    \answerNo{}
\end{enumerate}

\end{enumerate}

\appendix
% \onecolumn
\input{sections/7_supp}

\end{document}

%% file: sections/0_abstract.tex
Novel object captioning (NOC) aims to describe images containing objects without observing their ground truth captions during training. Due to the absence of caption annotation, captioning models cannot be directly optimized via sequence-to-sequence training or CIDEr optimization. As a result, we present \textit{Paraphrasing-to-Captioning (P2C)}, a two-stage learning framework for NOC, which would heuristically optimize the output captions via paraphrasing. With P2C, the captioning model first learns paraphrasing from a language model pre-trained on text-only corpus, allowing expansion of the word bank for improving linguistic fluency. To further enforce the output caption sufficiently describing the visual content of the input image, we perform self-paraphrasing for the captioning model with fidelity and adequacy objectives introduced. Since no ground truth captions are available for novel object images during training, our P2C leverages cross-modality (image-text) association modules to ensure the above caption characteristics can be properly preserved. In the experiments, we not only show that our P2C achieves state-of-the-art performances on nocaps and COCO Caption datasets, we also verify the effectiveness and flexibility of our learning framework by replacing language and cross-modality association models for NOC. Implementation details and code are available in the supplementary materials.

%% file: sections/1_intro.tex
\section{Introduction}
\label{sec:intro}

% Recently, the challenging task of image captioning is benefited from the progress in both computer vision~\citep{ren2015faster, anderson2018bottom, zhang2021vinvl} and natural language processing~\citep{vaswani2017attention, devlin2018bert, raffel2019exploring}. Incorporating the above technical advance, existing image captioning models~\citep{gao2019deliberate, huang2019attention, wang2019hierarchical, guo2020normalized, pan2020x, cornia2020meshed, zhou2020unified, li2020oscar, zhang2021vinvl} are able to accurately capture visual concepts in the image and translate them into beautiful sentences. Despite impressive benchmark performance on COCO Captions~\citep{chen2015microsoft} and Flickr~\citep{young2014image}, these models generalize poorly to images in the wild, since they are trained on datasets which only cover a tiny fraction of visual concepts in real world.
% \begin{figure*}[ht]
%   \centering
%   \label{teaser}
%   \includegraphics[page=5,trim={5 715 5 2}, clip, width=0.95\textwidth]{images/noc_architecture.pdf}
%   \vspace{-4mm}
%   \caption{Overview of VLAF2 captioning model. (a) Learning fluency by distilling knowledge of a paraphrase model. (b) Improving adequacy and fidelity of novel object captions via optimizing visual-linguistic association.}
%   \vspace{-5mm}
%   \label{Architecture}
% \end{figure*}

Novel Object Captioning (NOC)~\citep{agrawal2019nocaps} requires one to accurately describe images containing novel objects unseen during training captions. Despite impressive benchmark performance on COCO Captions~\citep{chen2015microsoft} and Flickr30K~\citep{young2014image}, existing image captioning models~\citep{huang2019attention, wang2019hierarchical, guo2020normalized, cornia2020meshed, zhou2020unified} or unsupervised image captioning works~\citep{gu2019unpaired, feng2019unsupervised} cannot generalize well. This is because existing unsupervised captioning models typically assume that training image and caption data share the same visual content (i.e., objects) of interest, which might not be held in practice~\citep{gu2019unpaired, feng2019unsupervised}.

Without observing captions describing novel objects during training, a number of NOC works choose to rely on object detection results for filling in the generated slotted sentences~\citep{lu2018neural, wu2018decoupled}. Since the object words and the template sentences are generated separately, their relationships might not be well described. Therefore,~\citet{hu2020vivo} propose to relate images and the produced captions by aligning the region features of an object and its associated word embedding, aiming at improved description of novel objects in scenes which are similar to the training ones. However, if a word is not presented in the training corpus, it might not be properly presented in the predicted caption. That is, the word embedding related to novel objects such as verbs and adjectives might not be fully exploited during inference, resulting in unsatisfactory \textit{linguistic fluency} of output captions. On the other hand, despite the visual feature of a novel object is aligned with textual feature, existing methods generally are not designed to sufficiently caption images containing such objects of interest. In other words, the \textit{fidelity} and \textit{adequacy} of the output captions cannot be preserved.

% \frank{While derivation of image-text aligned representation allows description of novel objects in similar scenes}, the associated wording such as verbs and adjectives might not be fully exploited during the captioning process. For example, for the lower caption in Fig.~\ref{fig:stage1}, one would prefer to use the phrase \textit{``wind through"} over \textit{``go through"}. On the other hand, despite the visual feature of object is aligned with textual feature, current methods are not explicitly enforced to caption the objects of interest. Take the lower branch in Fig.~\ref{fig:stage2} for example, the greedy-decoded (argmax) caption wrongly predicts the object \textit{``red scarf"} as a \textit{``tie"}. In addition to the correctness of the caption, a well-designed NOC model is expected to describe the scene thoroughly, describing as much visual gist as possible. For instance, for the upper caption in Fig.~\ref{fig:stage2}, it not only mentions the object \textit{``laptops"}, but also capture the object \textit{``white wires"}. The above deficiencies can be attributed to the fact that, these wordings, either a novel object or its associated verb, are not presented in the training corpus, thus they have low probability to be decoded during test time. 

In this paper, we uniquely approach the task of NOC by introducing and learning paraphrasing capabilities into state-of-the-art captioning models. More specifically, we advance pre-trained language models to expand the word bank of a captioning model for NOC, followed by enforcing the self-paraphrasing ability of this NOC model. The goal of the former stage is to preserve the linguistic fluency for NOC models, while that for the latter stage is deployed to exhibit improved fidelity and adequacy of the learned model. Since no ground truth captions of novel object images are available during training, we apply cross-modality association model with objectives/critics particularly designed for NOC.

%to heuristically optimize the resulting caption, together with two critics which function adversarially, a cross-modal association module and a novel repetition penalty module. The former is deployed to guarantee the paraphrased captions show improved fidelity and adequacy. While the latter is imposed to preserve the caption fluency during self-paraphrasing.

It is worth noting that, for the evaluation part of this work, we not only show that our method achieves state-of-the-art performances on the nocaps and COCO Caption datasets, we further assess the metrics reflecting the fluency, fidelity, and adequacy of output captions. In addition, via ablation studies, we further confirm the practicality and flexibility of our learning framework, which does not limit to particular language or cross-modality association modules for paraphrasing and image-text alignment.

%% file: sections/5_related.tex
\vspace{-2mm}
\section{Related Work}
\label{sec:related}
\textbf{Image captioning.}
Recent progress of image captioning focuses on different model architectures and learning methods.~\citet{ huang2019attention, wang2019hierarchical, guo2020normalized, cornia2020meshed} design different attention mechanisms for image captioning.~\citet{rennie2017self, li2019meta, yang2020fashion} adopt reinforcement learning to improve the performance. On the other hand, some researchers consider more challenging settings, such as partially supervised~\citep{liu2018show, kim2019image} or unpaired image captioning~\citep{gu2019unpaired, feng2019unsupervised}. However, these methods are restricted to the assumption that the unpaired images and captions share the same set of object class, and the number of object class is limited as well, which make them inapplicable to our task.

% ~\citet{zhou2020unified, li2020oscar, zhang2021vinvl, yang2021tap} conduct large-scale visual-language pretraining and has significant improvement after finetuning to the task of image captioning. However, these methods require a large amount of image-caption pairs, which is not applicable to our task.  

\textbf{Novel object captioning.}
Previously, novel object captioning approaches~\citep{anderson2016guided, anderson2018partially, hendricks2016deep} were only tested on a restrictive dataset with only eight novel object classes held out from the COCO dataset. Their extensions to large-scale image data with various novel objects are not sufficiently studied. Recent studies mainly rely on object detection results to improve the performance on novel object captioning.~\citet{lu2018neural, wu2018decoupled} generate slotted caption templates, which are later filled in with visual concepts identified by object detectors.~\citet{yao2017incorporating} exploit a copying mechanism to assemble words corresponding to object detector predictions to generate captions. Similarly, Constrained Beam Search (CBS)~\citep{anderson2016guided} is an architecture-agnostic decoding algorithm that can be exploited during inference to enforce the inclusion of novel object classes in the captions. Instead of explicitly using detection results,~\citet{hu2020vivo} and~\citet{vo2022noc} learn the relationship between image and text by aligning object detection tags with their corresponding image region features. Recently,~\citet{wang2021faier} indicate that a desirable caption should comprise properties of fluency, fidelity, and adequacy. Nevertheless, most existing NOC approaches are not designed to handle language expression and cross-modal association with the above properties preserved.

\textbf{Large-scale Vision and Language Pre-training (VLP).}
Recently, researchers discover that scaling up the sizes of both captioning model and training dataset would be effective to improve the performance on vision and language tasks. Dual-encoder frameworks such as CLIP~\citep{radford2021learning} and ALIGN~\citep{jia2021scaling} scale-up contrastive pre-training~\citep{oord2018representation} using 400M and 1.8B image-text pairs for cross-modal alignment. On the other hand, Transformer-based models like~\citep{li2020oscar, zhang2021vinvl} not only scale up the training corpus to 5.65M image and caption pairs, but also increase the transformer layers from 12 to 24. More recently, SimVLM~\citep{wang2021simvlm} and LEMON~\citep{hu2021scaling} further explore the \textit{huge} version of Transformer with a total of 32 layers, and scale up the pre-training corpus with 1.8B and 200M image-text pairs, respectively. Despite the dataset scale as an important factor in image captioning, we will demonstrate that our model still performs favorably against current large-scale methods under the same model size (i.e., Transformer~\textit{base} version, 12 layers) and with a smaller training caption corpus.

% Existing VLP works can be classified into two streams. One is based on the dual-encoder architecture, such as CLIP~\citep{radford2021learning} and ALIGN~\citep{jia2021scaling}, utilizing features encoded in each modality followed by contrastive learning~\citep{oord2018representation} for alignment purposes. On the other hand, models like~\citet{zhou2020unified, li2020oscar, zhang2021vinvl} exploit multiple cross-attention layers to learn the relationship between images and text, and show impressive performances on image-text matching tasks. While the former models are more efficient to handle data across modalities, the later models are more capable of performing fine-grained attention on cross-modal data.
% Due to the efficiency of the former models in handling data across modalities, we adopt CLIP for cross-modal association in this paper.

%% file: sections/3_method.tex
\begin{figure*}[tp]
  \centering
  \includegraphics[page=1,trim={5 580 5 3}, clip, width=0.95\textwidth]{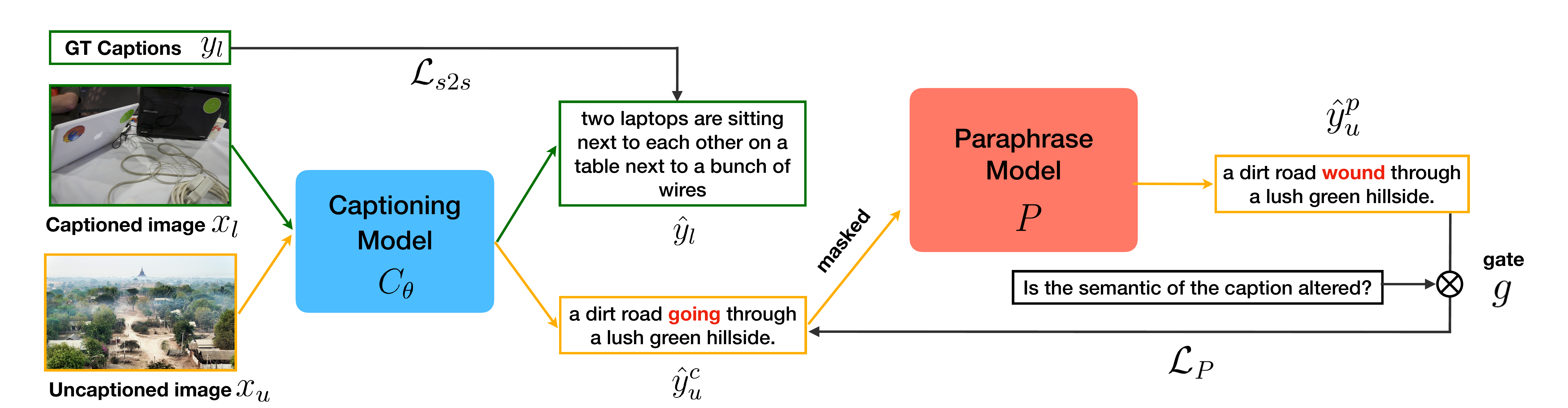}
  \vspace{-2mm}
  \caption{Learning to caption novel objects with linguistic fluency. For caption-labeled image $x_l$, we impose the sequence-to-sequence objective $\mathcal{L}_{s2s}$ for training. For uncaptioned image $x_u$, we exploit $P$ to paraphrase the generated caption $\hat{y}^c_u$, producing the refined caption $\hat{y}^p_u$, followed by a semantic-preserving gate $g$ to verify whether the paraphrased caption has altered the visual content.}
  \vspace{-2mm}
  \label{fig:stage1}
\end{figure*}

\section{Method}
\label{sec:method}

% By performing object-detection like training~\citep{carion2020end}, VIVO pretraining~\citep{hu2020vivo} enables our captioning model to predict novel object classes during inference. Still, the captioning model itself is not rewarded when generating captions with precise objects and beautiful wordings, or receives penalty when captions containing the wrong objects or general, uninformative descriptions. 
We first determine the notations and settings for the sake of completeness. Given a small set of images $X_l$ with the corresponding captions $Y_l$, as well as a large set of uncaptioned images $X_u$ containing novel objects, our goal is to generate the captions $\hat{Y}_u$ for $X_u$ via a captioning model $C_\theta$ ($\theta$ describes the parameters for the captioning model $C$). To achieve this, we propose \textit{Paraphrasing-to-Captioning (P2C)}, which allows $C$ to perform paraphrasing to preserve linguistic fluency, and to self-paraphrase for boosting the associated fidelity and adequacy. The former is regularized by a pre-trained language model $P$, while an image-text cross-modality model $A$ is utilized for guiding the unsupervised learning process.
%Since our framework is flexible, either $P$ or $A$ can be easily replaced with different implementation of language models or association models. 
We note that the contribution of our proposed P2C lies in how to leverage the paraphrasing mechanism, together with the linguistic and visual information from pre-trained models $P$ and $A$, for performing NOC with sufficient caption fluency, fidelity and adequacy.

%Take the lower branch in Fig.~\ref{fig:stage1} for example, when the captioning model generates a caption for a given image, our P2C would regularize and substitute the verb \textit{wound} for \textit{going}, improving the corresponding linguistic fluency. Followed by the learning stage in which we perform self-paraphrasing on the same image with two critics properly designed,~\frank{Title} is trained to produce captions with improved fidelity and adequacy. For example, in Fig.~\ref{fig:stage2}, the caption in the lower box accurately describes the object \textit{red scarf} and thus receives a higher score from $A$. This encourages the model to correctly describe the visual content and sufficiently capture the associated visual concept of the input image. In the following subsections, we will detail how we exploit the paraphrasing mechanism, together with a language model $P$, an association model $A$, for improving fluency, fidelity, and adequacy for NOC.

%We discuss why we use BERT and CLIP to optimize for fluency, fidelity and adequacy in Sec.~\ref{subsec:connection}. Then, in Sec.~\ref{subsec:bert} and~\ref{subsec:clip}, we detail the various ingredients of our two-stage learning framework.

\subsection{Describing Novel Objects with Linguistic Fluency}
\label{subsec:bert}

By observing image-caption pairs $(x_l, y_l)$, the captioning model $C_\theta$ in Fig.~\ref{fig:stage1} would learn the visual grounding (i.e., localization of known objects and referring their expressions) as well the linguistics information. The training objective of such labelled data is a conventional sequence-to-sequence loss $\mathcal{L}_{s2s}$, calculated on ground-truth caption $y_l$. That is, we have $\mathcal{L}_{s2s} = \text{CrossEntropy}(\hat{y}_l, y_l)$.

However, using the image-caption pairs $(x_l, y_l)$ alone is not sufficient to produce fluent captions for uncaptioned images $X_u$. We observe that some commonly-used wording of the associated novel objects, are not presented in the training corpus. Therefore, we leverage a language model $P$, to learn its linguistic knowledge via \textit{paraphrasing}. That is, for uncaptioned images $X_u$ containing novel objects, we first generate a caption describing $X_u$. Then, we leverage pre-trained language models as the paraphrase model $P$, which replace the generated captions with the most probable wording of a given novel object context. A semantic-preserving gate $g$ is deployed to validate the paraphrased caption. As detailed later, this gate $g$ is realized by a cross-modal association model $A$, which assesses the relationship between the paraphrased caption and the input image, ensuring semantics of the caption is not modified by $P$.

\textbf{Learning to paraphrase via language model $P$.} We now present the detailed paraphrasing processing for learning $C_\theta$. As illustrated in Fig.~\ref{fig:stage1}, given an uncaptioned image $x_u$, the captioning model $C_\theta$ generates a caption $\hat{y}_u^c = C_\theta(x_u)$, $\hat{y}_u^c = \{w^c_1, w^c_2, ..., w^c_T\}$, where $w^c_i$ denotes the $i$th word, and $T$ is the caption length. The superscript $c$ represents it is generated by our captioning model. We then obtain a masked caption $\hat{y}_u^M= \{w^c_1, w^c_2, ..., w^M_m, ..., w^c_T\}$ with $m$ indicating the mask index, by randomly masking out the words in the caption. We note that, we do \textit{not} mask nouns/objects in the above process, since they are related to objects grounded in the (novel) visual content and thus are not explicitly associated with caption fluency. As a result, $P$ takes the masked sequence $\hat{y}_u^M$ as input and predicts the masked word $w_m^p$ with the highest probability conditioned on the context of the entire sentence, producing the paraphrased caption $\hat{y}_u^p = P(\hat{y}_u^M)$, $\hat{y}_u^p= \{w^c_1, w^c_2, ..., w^p_m, ..., w^c_T\}$.

\textbf{Semantic-preserving gate $g$.} In the above paraphrasing process, however, not every word substitution from $P$ is guaranteed to be semantically correct. We thus require a proper image-text model (i.e., association model) $A$ to validate the paraphrasing output. That is, if the paraphrased caption comprises a more accurate and associated word that human generally uses to describe the scene, then a higher score would be obtained (than that of the original caption). Thus, we propose the objective function $\mathcal{L}_{P}$ to calculate the loss for the replaced words, gated by the comparison of the association scores of the corresponding captions. More precisely, $\mathcal{L}_{P}$ is derived as:
\begin{equation}
    \label{bert_eq}
    \begin{aligned}
        \mathcal{L}_{P} = - g * t\log(s), \quad 
        g = \max(0, \tanh(A(x_u, y_u^p) - A(x_u, y_u^c) - \alpha))
        % \begin{cases}
        %     0 & \text{if } A(x_u, y_u^p) \leq A(x_u, y_u^c) + \alpha\\
        %     1 & \text{if } A(x_u, y_u^p) > A(x_u, y_u^c) + \alpha
        % \end{cases},
    \end{aligned}
\end{equation}
where $g$ is the gating function preventing $C_\theta$ from learning from low-quality paraphrased captions, $t$ is the one-hot representation of the paraphrased word, and $s$ is the word distribution predicted by the captioning model at the masked timestep $m$, i.e., we have $w^c_m = \arg\max(s)$. We have $A(x,y)$ in $g$ to calculate the association between an image $x$ and its caption $y$, indicating how well the captions match the images, and $\alpha$ serves as the margin of the association scores for distinguishing between the two captions.

\begin{figure*}[tp]
  \centering
  \includegraphics[page=2,trim={5 600 5 5}, clip, width=0.95\textwidth]{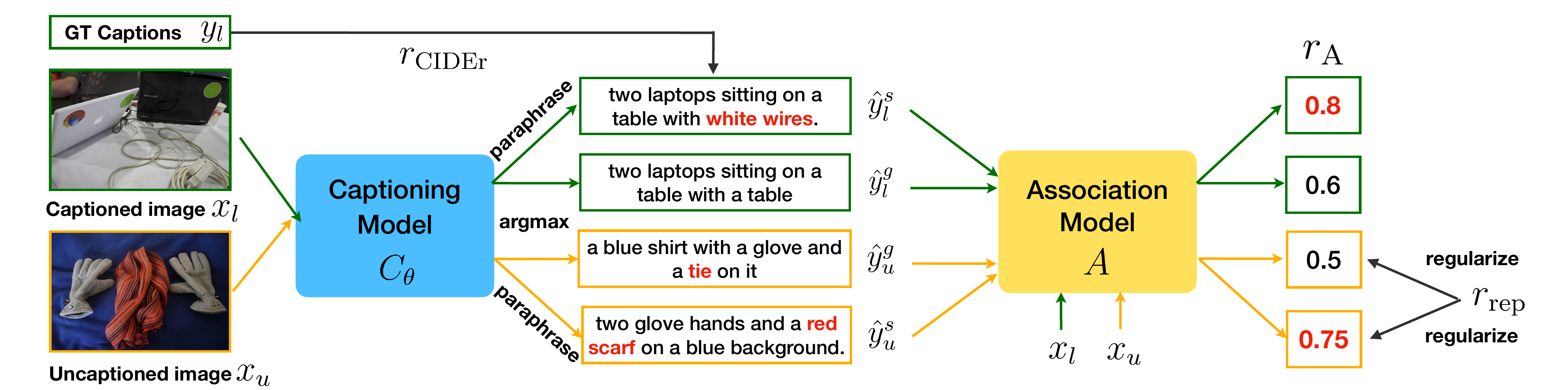}
  \vspace{-2mm}
  \caption{Improving caption fidelity and adequacy via self-paraphrasing. For caption-labeled image $x_l$, we perform CIDEr optimization. The sampled caption $\hat{y}_d^s$ will be rewarded by $A$ if it has a higher cross-modal association than the greedy-decoded baseline $\hat{y}_d^g$. The superscript $d$ indicates the source of the image. Additionally, we regularize our model with $r_\text{rep}$ to avoid repetitive captions.}
  \vspace{-5mm}
  \label{fig:stage2}
\end{figure*}

\subsection{Improving Caption Fidelity and Adequacy via Self-Paraphrasing}
\label{subsec:clip}

Recall that, as discussed in Sec.~\ref{sec:intro}, caption fidelity verifies whether the details of visual content are presented in the generated caption, while adequacy assesses whether the visual content is properly described  been expressed in it. While the previous paraphrasing stage increases the linguistic fluency for captions describing images containing novel objects, the novel object word itself still can be incorrectly predicted or missing in the output captions. This is because that, the caption corpus $Y_l$ does not contain any novel objects, and thus the novel object words have low probability to be captioned. 

To tackle the above issues, one possible solution is to \textit{self-paraphrasing} for $C_\theta$ with captioning evaluation metric as a critic to reward the paraphrased captions. For labeled images $X_l$, we can exploit the CIDEr score~\citep{rennie2017self}, encouraging the generated caption to be consistent with that of the human annotated ones in the word level. 
% Conventional sequence-to-sequence model training with cross-entropy loss might not reflect the above desirable properties. This is because that, conventional sequence-to-sequence objective simply calculates the average word-level accuracy of the generated sentences, which is unable to assess whether the caption accurately and sufficiently describes particular novel objects. To tackle the above issues, one possible solution is to utilize self-paraphrasing and take the captioning evaluation metric as a critic to evaluate and heuristically optimize the resulting caption. For labeled images $X_l$, we can exploit the CIDEr score~\citep{rennie2017self}, encouraging the generated caption to be consistent with that of the human annotated ones in the word level. 
However, while CIDEr can be easily computed for captioning labeled images $X_l$, it cannot be explicitly calculated for captioning uncaptioned images $X_u$ due to the absence of ground-truth captions. To address this problem, we discover that cross-modal association is an ideal measurement to reflect the fidelity and adequacy for output captions (see remarks in Appendix~\ref{appendix:connection}). Specifically, we utilize an association model $A$ to compute the association score between images and the generated captions, which serves as a \textit{label-free} critic in our learning framework. Then, when a paraphrased caption is rewarded by a higher association score, our P2C would increase the probability of using that word in the sentence to describe the image. This learning strategy allows producing captions that precisely describe the objects with plentiful visual details. 
% Due to page limits, fundamental supports for connecting cross-modal association to both fidelity and adequacy are detailed in Appendix~\ref{appendix:connection}.

\textbf{Rewarding the generated captions.} 
As a potential challenge in NOC, we observe that captioning models would achieve improved association by simply repeating the same object that occurs in the image, which undermines the linguistic fluency of the captions. For example, the image caption \textit{``a group of cans of soda and other items on a table"} can be replaced by \textit{``a pile of \textbf{cans} and bottles of soda on a counter with \textbf{cans} of \textbf{cans}"} with a higher association score. 

To tackle this problem, we choose to impose a repetition penalty to avoid such trivial solutions. For image-caption pairs $(x_l, y_l)$, we directly calculate the CIDEr reward for the predicted caption $\hat{y}_l$ (i.e., $r_\text{CIDEr} = \text{CIDEr}(\hat{y}_l, y_l)$). To enforce the generated captions for $(X, Y) = (X_l, Y_l) \cup (X_u, Y_u)$ with sufficient fidelity and adequacy, we exploit $A$ to compute the association reward between $X$ and $\hat{Y}$ (i.e., $r_\text{A} = A(x, \hat{y})$). As for the repetition penalty to preserve linguistic fluency of the generated captions $\hat{y}_u = \{w1, w2, ..., w_T\}$ for $x_u$, we formulate it as a linear assignment problem, where every word is assigned to the most similar one in the same sentence except for itself. Then, we calculate the similarity between such pairs for each sentence. Intuitively, repetitive captions would have high similarity scores (with repeating words assigned to the exact same word)s. Thus, we define the assignment $\hat{\alpha}$ as the one that maximizes the average pairwise similarity of a sentence, i.e.,
\begin{equation}
    \label{rep_pen}
    \begin{aligned}
        \hat{\alpha} = \mathop{\arg\max}_{\alpha} \frac{1}{T} \sum_{i=1}^T C(w_i, w_{\alpha(i)}), 
    \end{aligned}
\end{equation}
where $\alpha(i)$ is the index of the word assigned to the $i$-th word in the caption, and $C(w_i, w_j)$ is the cosine similarity between the GloVe~\citep{pennington2014glove} word representation of two words. Since a desirable captioning model would encourage captions with low repetition (i.e., low average pairwise similarity), the reward for repetition penalty is defined as follows:
\begin{equation}
    \label{rep_reward}
    \begin{aligned}
        r_\text{rep} = 1 - \frac{1}{T} \sum_{i=1}^T C(w_i, w_{\hat{\alpha}(i)}), 
    \end{aligned}
\end{equation}
Note that we do not calculate repetition penalty for the labeled data $\hat{Y}_l$ since they are regularized by the aforementioned CIDEr rewards. 

With the above discussions, the total reward for caption-labeled data would be $r(\hat{y}_l) = r_\text{CIDEr}(\hat{y}_l, y_l) + r_\text{A}(x_l, \hat{y}_l)$, and the total reward for uncaptioned data would be $r(\hat{y}_u) = r_\text{A}(x_u, \hat{y}_u) + r_\text{rep}(\hat{y}_u)$.

\textbf{Back-propagation via reinforce algorithm.} 
Unfortunately, computation of the aforementioned rewards is non-differentiable. Thus, we adopt reinforce algorithm~\citep{williams1992simple} to optimize our P2C learning. As shown in Fig.~\ref{fig:stage2}, for an image $x$ we use greedy decoding to obtain the baseline result $\hat{y}^g$, and the paraphrased captions $\hat{y}^s$ are derived from randomly sampling from the word distribution. If the sampled captions possess higher linguistic fluency or cross-modal association than the baseline caption, they will be encouraged by positive rewards and vice versa. We follow~\citet{rennie2017self, liu2017improved, liu2018show} and define the objective function as follows:
\begin{equation}
    \label{rl}
    \begin{aligned}
        \nabla_{\theta}\mathcal{L}_{RL}(\theta) \approx -(r(\hat{y}^s_d)-r(\hat{y}^g_d)) \nabla_{\theta} \log p_{\theta}(\hat{y}^s_d), \\
        r(\hat{y}_d) = \begin{cases}
             r_\text{CIDEr}(\hat{y}_d, y_d) + r_\text{A}(x_d, \hat{y}_d) & \text{if } x_d \in X_l \\
            r_\text{A}(x_d, \hat{y}_d) + r_\text{rep}(\hat{y}_d) & \text{if } x_d \in X_u
        \end{cases},
    \end{aligned}
\end{equation}
where $d$ indicates the source of the image, $\theta$ being the parameters of captioning model, and $p_{\theta}(\hat{y}^s)$ represents the predicted word logits for the generated captions. With the objective functions defined in equations (\ref{bert_eq}), and (\ref{rl}), the NOC model $C_\theta$ can be trained accordingly.

% To directly optimize for more associated captions, we cast our captioning models in the Reinforcement Learning terminology following~\citet{rennie2017self, liu2017improved, liu2018show}. Our captioning model acts as an ``agent'', interacting with an ``environment'' (image feaures and the previous generated words). The parameters of the captioning model, $\theta$, define a policy $p_{\theta}$, which is the prediction logits of the next word, 

% \begin{table}[tp]
% \caption{Quantitative results on the nocaps (XD) test set.}
% \vspace{-1mm}
% \label{nocapsxd_table}
% % \renewcommand{\arraystretch}{1.00}
% % \begin{adjustbox}{max width=1.0\textwidth,center}
% \begin{tabular}{ccc}
% \hline
% \multicolumn{1}{l|}{\multirow{2}{*}{Method}} & \multicolumn{2}{c}{overall} \\ 
% \multicolumn{1}{c|}{} & \multicolumn{1}{c}{CIDEr} & SPICE \\ \hline
% \multicolumn{1}{l|}{UpDown~\citep{agrawal2019nocaps}} &  73.09 & 11.20 \\ 
% \multicolumn{1}{l|}{$\text{SimVLM}_\text{base}$~\citep{wang2021simvlm}} & 94.80 & 13.10 \\ 
% \multicolumn{1}{l|}{VIVO~\citep{hu2020vivo}} & 100.12 & 14.04 \\ \hline
% \multicolumn{1}{l|}{Ours} &  96.25 & 14.10 \\ 
% \multicolumn{1}{l|}{Ours (+Conceptual Captions)} &  \textbf{102.39} & \textbf{14.71} \\ \hline
% \end{tabular}
% % \end{adjustbox}
% \vspace{-6mm}
% \end{table}

%% file: sections/4_exp.tex
\section{Experiments}
\label{sec:exp}

\textbf{Implementation details.} Following~\citet{hu2020vivo,li2020oscar, zhang2021vinvl}, we consider a BERT-base~\citep{devlin2018bert} architecture for our captioning model $C_\theta$. To demonstrate the flexibility of our learning framework, we apply two different versions of BERT,~\textit{base} and ~\textit{large}, which are pre-trained on large-scale \textit{text}-only corpus to model human language as our paraphrase model $P$. For the association model $A$, we have three different cross-modal association models, VIFIDEL~\citep{madhyastha2019vifidel}, SR-PL~\citep{liu2018show}, and CLIP~\citep{radford2021learning} with its version being ViT/B-32. VIFIDEL associates image-caption data using word embedding of particular object labels, and SR-PL utilizes the triplet loss to learn the association of image captions, while CLIP is optimized via the contrastive pre-training. For our model reported in the following subsections, we use BERT large for $P$, and CLIP for $A$. Due to page limits, hyperparameters and other training details can be found in Appendix \ref{appendix:implementation}.

\textbf{Datasets.} The training data for the \textbf{nocaps} benchmark comprises the Open Images V4~\citep{kuznetsova2020open} object detection training set (1.7M images annotated with bounding boxes for 600 object classes), plus the image-caption pairs from the COCO Captions 2017~\citep{chen2015microsoft} training set (0.5M image-caption pairs containing 80 object classes). No additional image-caption pairs are provided for training. We evaluate our model on the validation and test set of nocaps, which comprises 4500 and 10600 images from the Open Images validation and test sets, respectively. To compare with current large-scale methods and investigate our model performance when scaling up the training dataset, we jointly trained our model on the Conceptual Caption dataset~\citep{sharma2018conceptual} and compare our method to other methods that access additional image-text pairs during their training on the nocaps-XD~\citep{agrawal2019nocaps} benchmark (XD stands for extra data). In addition, to demonstrate that our method also enhances model performance on the general image captioning, we evaluate our P2C on the COCO Captions dataset. Due to page limits, the comparison on COCO Captions can be found in Appendix~\ref{appendix:cocoexp}.

\begin{table*}[t]
\caption{Quantitative results on nocaps. Note that C and S denote CIDEr and SPICE, respectively. We highlight the highest score in blue, while the second best scores are marked in bold.}
\vspace{1mm}
\label{quantitative_table}
\begin{adjustbox}{max width=1.0\textwidth, center}
\begin{tabular}{l||cc|cc|cc|cc||cc|cc|cc|cc}
\hline
\multirow{2}{*}{Method} & \multicolumn{2}{c|}{in-domain} & \multicolumn{2}{c|}{near-domain} & \multicolumn{2}{c|}{out-domain} & \multicolumn{2}{c||}{overall} & \multicolumn{2}{c|}{in-domain} & \multicolumn{2}{c|}{near-domain} & \multicolumn{2}{c|}{out-domain} & \multicolumn{2}{c}{overall} \\
 & C & S & C & S & C & S & C & S & C & S & C & S & C & S & C & S\\ \hline
 & \multicolumn{8}{c||}{Validation Set} & \multicolumn{8}{c}{Test Set} \\ \hline
UpDown & 79.3 & 12.4 & 73.8 & 11.4 & 71.7 & 9.9 & 74.3 & 11.2 & 76.0 & 11.8 & 74.2 & 11.5 & 66.7 & 9.7 & 73.1 & 11.2 \\
Oscar$_B$ & 83.4 & 12.0 & 81.6 & 12.0 & 77.6 & 10.6 & 81.1 & 11.7 & 81.3 & 11.9 & 79.6 & 11.9 & 73.6 & 10.6 & 78.8 & 11.7 \\
Oscar$_L$ & 85.4 & 11.9 & 84.0 & 11.7 & 80.3 & 10.0 & 83.4 & 11.4 & 84.8 & 12.1 & 82.1 & 11.5 & 73.8 & 9.7 & 80.9 & 11.3 \\
\begin{tabular}[c]{@{}l@{}}Oscar$_B$\\ \ \ +VIVO\end{tabular} & 92.2 & 12.9 & 87.8 & 12.6 & 87.5 & 11.5 & 88.3 & 12.4 & 89.0 & 12.9 & 87.8 & 12.6 & 80.1 & 11.1 & 86.6 & 12.4 \\
VinVL & \textbf{96.8} & 13.5 & 90.7 & 13.1 & 87.4 & 11.6 & 90.9 & 12.8 & 93.8 & 13.3 & 89.0 & 12.7 & 66.1 & 10.9 & 85.5 & 12.5 \\
\begin{tabular}[c]{@{}l@{}}VinVL\\ \ \ +VIVO\end{tabular} & 94.8 & 13.3 & \textbf{91.4} & 13.0 & 88.7 & 11.6 & \textbf{91.4} & 12.7 & \textbf{94.5} & 13.1 & \textbf{90.9} & 12.9 & 77.9 & 11.3 & \textbf{88.3} & 12.6\\ \hline
Human & 84.8 & \textbf{14.3} & 85.0 & \textbf{14.3} & \highest{95.7} & \textbf{14.0} & 87.1 & \highest{14.2} & 80.6 & \highest{15.0} & 84.6 & \highest{14.7} & \highest{91.6} & \highest{14.2} & 85.3 & \highest{14.6}\\ \hline
Ours & \highest{101.4} & \highest{15.1} & \highest{96.8} & \highest{14.5} & \textbf{95.4} & \textbf{12.9} & \highest{97.2} & \highest{14.2} & \highest{101.7} & \highest{15.0} & \highest{95.7} & \textbf{14.4} & \textbf{82.5} & \textbf{12.2} & \highest{93.5} & \textbf{14.1}\\ \hline
\end{tabular}
\end{adjustbox}
\vspace{-4mm}
\end{table*}

\vspace{-2mm}
\subsection{Evaluation metrics}
\label{subsec:metrics}

\textbf{CIDEr.} Like the evaluation metrics~\citep{papineni2002bleu, lin2004rouge, banerjee2005meteor} in NLP, Consensus-based Image Description Evaluation (CIDEr) calculates the similarity between the reference and generated caption by n-gram overlap in a rule-based manner. To capture human consensus in image captioning, it introduces the tf-idf weight to reduce the matching weight of the n-grams that are common in all image captions.

\textbf{SPICE.} Semantic Propositional Image Caption Evaluation (SPICE)~\citep{anderson2016spice} matches the semantics between sentences, such as objects, relations, and attributes of objects. Specifically, it converts sentences into semantic scene graphs, which allows evaluation to break grammatical constraints and focuses on propositional semantic content. It reflects the accuracy of the visual content and considers less about linguistic properties.

\textbf{Fluency.} To quantitatively evaluate fluency, we remove the effect of the visual information and focus on the quality of linguistic properties in the conventional caption evaluation metrics. Specifically, we we disregard the n-grams containing the particular object word of interest during the computation BLEU@4~\citep{papineni2002bleu} and CIDEr scores. Take the sequence “a b c d” for example, when ‘b’ is the object word, only the unigram “a, c, d” and the bigram “cd” would be taken into account, n-grams such as “ab” or “abc” would be excluded from computation. Note that the fluency experiment is conducted on a subset of the nocaps validation set, which contains 1000 images whose caption annotations are available on the official website of the nocaps dataset. 

\textbf{Fidelity \& Adequacy.} Fidelity and adequacy evaluate how well the captions are associated with images. As defined in Sec.~\ref{sec:intro}, fidelity evaluates whether the objects described by the caption are actually presented in the images, while adequacy assesses how many objects in the images are described by the captions. These two properties are analogous to the definition of precision and recall, respectively. Therefore, we extract the objects mentioned in the captions and the ground-truth objects in the images and calculate the precision (for fidelity), recall (for adequacy), and F1 (for overall association) scores. The experiment is performed on the validation set of nocaps.

Following~\citet{agrawal2019nocaps}, we split the dataset into three subsets for evaluation: \textit{in-domain} images only contain the seen objects that have been described during training, \textit{out-of-domain} images are the unseen/uncaptioned (i.e., novel) ones, while \textit{near-domain} ones contain both seen and unseen objects.

\vspace{-2mm}
\subsection{Quantitative analysis}

For performance comparisons, we choose UpDown~\citep{agrawal2019nocaps} as baselines, as well as Oscar~\citep{li2020oscar} and VinVL~\citep{zhang2021vinvl} which achieves SOTA results on the benchmark of nocaps. All methods are trained via the SCST optimization~\citep{rennie2017self} except for the UpDown baseline, and Constrained Beam Search (CBS) is exploited during inference. In addition, VIVO~\citep{hu2020vivo} is a pre-training technique for captioning models, allowing them to recognize the novel objects. Since VinVL did not report the numbers with CBS exploited during inference (CBS is known to improve model performance on out-of-domain data), we reproduce VinVL following details stated in the original paper. For more details please refer to Appendix~\ref{appendix:implementation}. For a comprehensive comparison, we conduct experiments on the validation and test set of nocaps. In addition, we compare our method with VinVL and VIVO in fluency, fidelity, and adequacy to demonstrate the improvement in terms of these properties. We also evaluate on the COCO Caption dataset and report in Appendix~\ref{appendix:cocoexp}.

\textbf{The nocaps datasets.} The results on nocaps are shown in Table~\ref{quantitative_table}. From this table, we see that our model performed favorably against baselines and SOTAs across different metrics. 
% We observe that CBS slightly decreased model performance on \textit{seen} object captioning, since it forces captioning models to describe the \textit{detected} objects without considering detection correctness. Nevertheless, for near or out-of-domain images, CBS still benefits the captioning performances. 
We see that our method substantially improves the CIDEr scores in every domain, which verifies our design to generate  more fluent and natural captions. Also, it is worth noting that our model largely increased the performance in SPICE score for every data domain, which verifies that our method is able to generate captions with the improved image-language association. 

\textbf{The nocaps-XD datasets.} To compare with large-scale pre-training methods and investigate our model performance when scaling up training dataset, we evaluate our method on the nocaps-XD benchmark and show the results in Table~\ref{nocapsxd_table}. Note that both VinVL, SimVLM, LEMON, and our method adopt BERT-based backbones for captioning. 
% We note that, VIVO did not specify the details of datasets additionally applied, while SimVLM specifically pointed out that a much larger web-scale dataset (1.8B image-text pairs) was utilized comparing to our use of Conceptual Caption dataset (3.1M image-text pairs). 
From the above table, we see that our method still performs favorably against SOTAs on the benchmark even if we have the fewest training caption samples (3M). Our model performance is only slightly below LEMON on the CIDEr score on validation set when it uses a larger training corpus (ALT200M), while we surpass LEMON by a significant margin when the same training corpus (CC3M) is used. Thus, the effectiveness of our method when we scale up the training data can be verified. For more detailed discussion and the ablation study on this model, please refer to Appendix~\ref{appendix:nocapsxd}.

% \begin{table}[t]
% \caption{Analysis on the association model $\mathit{A}$.}
% \vspace{0mm}
% \label{ablation_association_table}
% \setlength{\tabcolsep}{16pt}
% \renewcommand{\arraystretch}{1.1}
% \begin{adjustbox}{max width=1.0\textwidth, center}
% \setlength{\tabcolsep}{6mm}{
% \begin{tabular}{ccc}
% \hline
% \multicolumn{1}{l|}{\multirow{2}{*}{Method}} & \multicolumn{2}{c}{overall} \\ 
% \multicolumn{1}{c|}{} & \multicolumn{1}{c}{CIDEr} & SPICE \\ \hline
% \multicolumn{1}{l|}{Ours w/ VIFIDEL} & 55.32 & 10.73 \\ 
% \multicolumn{1}{l|}{Ours w/ DISC} & 84.58 & 13.25 \\ 
% \hline
% \multicolumn{1}{l|}{Ours w/ CLIP} & \textbf{96.25} & \textbf{14.10} \\ 
% \hline
% \end{tabular}}
% \end{adjustbox}
% \vspace{-4mm}
% \end{table}

\textbf{Fluency, fidelity, and adequacy.} 
As described in Sec.~\ref{subsec:metrics}, we design additional experiments for evaluating fluency, fidelity, and adequacy and report the results in Table~\ref{metrics_1}. For fluency, we remove all the objects and nouns in the captions since they relate less to the linguistics of the captions. We then calculate the BLEU@4 (B@4) and CIDEr (C) scores for the captions after removal. For fidelity and adequacy, they indicate that captions should accurately (high precision) describe sufficient (high recall) visual details. Therefore, we report the object precision and recall in this table, and object F1 scores represent the overall association between captions and images. One can see that our method surpasses previous methods by a visible margin on all tasks except for in-domain object precision, which further verifies our model improves novel object captioning on fluency, fidelity, and adequacy.

\begin{table*}[tp]
% \small
\caption{Quantitative results on the nocaps (XD) benchmark. Note that XD stands for extra data.}
\vspace{1mm}
\label{nocapsxd_table}
\setlength{\tabcolsep}{14pt}
\renewcommand{\arraystretch}{1.0}
\begin{adjustbox}{max width=\textwidth, center}
\begin{tabular}{l|l|cc|cc}
\hline
\multirow{2}{*}{Method} & \multirow{2}{*}{Pre-training data} & \multicolumn{2}{c|}{Validation set} & \multicolumn{2}{c}{Test set} \\
 &  & CIDEr & SPICE & CIDEr & SPICE \\ \hline
Encoder-Decoder~\citep{changpinyo2021conceptual} & CC12M~\citep{changpinyo2021conceptual} & 87.4 & 11.8 & 85.3 & 11.8 \\
Encoder-Decoder & CC3M+CC12M & 90.2 & 12.1 & 87.3 & 12.0 \\ 
VinVL$_{base}$~\citep{zhang2021vinvl} & 5.65M Combined & 95.5 & 13.5 & - & - \\
SimVLM$_{base}$~\citep{wang2021simvlm} & 1.8B & - & - & \textbf{94.8} & 13.1 \\
LEMON$_{base}$~\citep{hu2021scaling} & CC3M~\citep{sharma2018conceptual} & 91.6 & 13.0 & - & - \\
LEMON$_{base}$ & CC12M & 100.4 & 13.8 & - & - \\
LEMON$_{base}$ & ALT200M~\citep{hu2021scaling} & \highest{106.8} & 14.1 & - & - \\ \hline
Ours & COCO Caption 0.5M & 97.2 & \textbf{14.2} & 93.5 & \textbf{14.1} \\
Ours & CC3M & \textbf{104.1} & \highest{14.6} & \highest{102.4} & \highest{14.7} \\ \hline
\end{tabular}
\end{adjustbox}
\end{table*}

\begin{table*}[tp]
\caption{Quantitative comparisons on caption fluency, fidelity and adequacy. Note that BLEU@4 (B@4) and CIDEr (C) are utilized for describing fluency, object precision (P) for fidelity, object recall (R) for adequacy and object F1 scores (F1) for overall cross-modal association.}
\vspace{2mm}
\label{metrics_1}
\renewcommand{\arraystretch}{1.1}
\begin{adjustbox}{max width=1.0\textwidth,center}
\begin{tabular}{l|ccccc|ccccc|ccccc}
\hline
\multirow{2}{*}{Method} & \multicolumn{5}{c|}{in-domain} & \multicolumn{5}{c|}{near-domain} & \multicolumn{5}{c}{out-of-domain} \\
% & BLEU@4 & CIDEr & Fidelity & Adequacy & F1 & BLEU@4 & CIDEr & Fidelity & Adequacy & F1 & BLEU@4 & CIDEr & Fidelity & Adequacy & F1 \\ \hline
 & B@4 & C & P & R & F1 & B@4 & C & P & R & F1 & B@4 & C & P & R & F1 \\ \hline
VinVL & 21.6 & 74.8 & \textbf{59.2} & 40.8 & 48.3 & 19.6 & 73.3 & 22.8 & 32.6 & 26.8 & 17.9 & 59.6 & 48.4 & 25.6 & 33.5 \\
VinVL+VIVO & 20.6 & 71.6 & 56.0 & 42.2 & 48.1 & 19.8 & 75.4 & 28.5 & 36.3 & 32.0 & 17.4 & 59.6 & 49.0 & 27.3 & 35.1 \\
\textbf{Ours} & \textbf{25.4} & \textbf{87.0} & 58.2 & \textbf{45.6} & \textbf{51.3} & \textbf{22.1} & \textbf{80.1} & \textbf{39.9} & \textbf{41.0} & \textbf{40.4} & \textbf{19.7} & \textbf{67.7} & \textbf{51.3} & \textbf{30.5} & \textbf{38.3} \\
\hline
\end{tabular}
\end{adjustbox}
\vspace{-4mm}
\end{table*}

\begin{table}[t]
% \small
\caption{Analyses on paraphrasing model $P$, association model $A$, and repetition penalty for NOC using nocaps validation set. Note that $P$ mainly benefits the linguistic fluency with improved CIDEr, and reward from $A$ is desirable for preserving visual semantics with increased SPICE.}
\vspace{1mm}
\label{ablation_table}
\setlength{\tabcolsep}{10pt}
\renewcommand{\arraystretch}{1.1}
\begin{adjustbox}{max width=1.0\textwidth,center}
\begin{tabular}{lcccccccc}
\hline
\multicolumn{1}{l|}{\multirow{2}{*}{Method}} & \multicolumn{2}{c|}{in-domain} & \multicolumn{2}{c|}{near-domain} & \multicolumn{2}{c|}{out-domain} & \multicolumn{2}{c}{overall} \\
\multicolumn{1}{c|}{} & \multicolumn{1}{c}{CIDEr} & \multicolumn{1}{c|}{SPICE} & \multicolumn{1}{c}{CIDEr} & \multicolumn{1}{c|}{SPICE} & \multicolumn{1}{c}{CIDEr} & \multicolumn{1}{c|}{SPICE} & \multicolumn{1}{c}{CIDEr} & \multicolumn{1}{c}{SPICE} \\ \hline
% \multicolumn{9}{c}{Validation Set} \\ \hline
\multicolumn{1}{l|}{\textbf{Ours}} & \multicolumn{1}{c}{\textbf{102.8}} & \multicolumn{1}{c|}{\textbf{14.8}} & \multicolumn{1}{c}{\textbf{97.9}} & \multicolumn{1}{c|}{\textbf{14.4}} & \multicolumn{1}{c}{\textbf{86.3}} & \multicolumn{1}{c|}{\textbf{12.5}} & \multicolumn{1}{c}{\textbf{96.3}} & \multicolumn{1}{c}{\textbf{14.1}} \\ \hline
\multicolumn{1}{l|}{Ours w/o $g$} & 32.8 & \multicolumn{1}{c|}{10.6} & 21.5 & \multicolumn{1}{c|}{9.5} & 12.6 & \multicolumn{1}{c|}{8.0} & 21.3 & 9.4 \\
\multicolumn{1}{l|}{Ours w/o $\mathcal{L}_{P}$} & 99.1 & \multicolumn{1}{c|}{14.4} & 94.7 & \multicolumn{1}{c|}{14.1} & 84.5 & \multicolumn{1}{c|}{12.4} & 93.3 & 13.8 \\
\multicolumn{1}{l|}{Ours w/o $r_\text{A}$} & 101.1 & \multicolumn{1}{c|}{13.8} & 94.1 & \multicolumn{1}{c|}{13.4} & 80.5 & \multicolumn{1}{c|}{11.9} & 92.3 & 13.1 \\ 
\multicolumn{1}{l|}{Ours w/o $r_\text{rep}$} & 96.7 & \multicolumn{1}{c|}{\textbf{14.8}} & 89.6 & \multicolumn{1}{c|}{14.1} & 81.9 & \multicolumn{1}{c|}{12.4} & 89.1 & 13.9 \\\hline
% \multicolumn{1}{l|}{Ours w/o rep. penalty} & 97.49 & \multicolumn{1}{c|}{\textbf{15.55}} & 91.05 & \multicolumn{1}{c|}{\textbf{15.16}} & 82.14 & \multicolumn{1}{c|}{\textbf{12.85}} & 90.17 & \textbf{14.76} \\ \hline
\end{tabular}
% }
\end{adjustbox}
\end{table}

% \begin{figure*}[tp]
%   \centering
%   \includegraphics[page=1,trim={200 10 190 20}, clip, width=\textwidth]{images/quantitative_2.pdf}
%   \vspace{-2mm}
%   \caption{Analyses on paraphrase model $P$ and association model $A$ for improving caption fluency, fidelity and adequacy. Note that $P$ benefits fluency metrics of BLEU@4 and CIDEr, while $A$ focusing on cross-modal association boosts metrics of object precision, recall, and F1 scores.}
%   \vspace{-1mm}
%   \label{fig:quantitative_2}
% \end{figure*}

\vspace{-2mm}
\begin{table*}[t]
\caption{Analyses on paraphrase model $P$ and association model $A$ for improving caption fluency, fidelity and adequacy. Note that $P$ benefits fluency metrics of BLEU@4 and CIDEr, while $A$ focusing on cross-modal association boosts metrics of object precision, recall, and F1 scores.}
\vspace{2mm}
\label{metric_2}
\renewcommand{\arraystretch}{1.1}
\begin{adjustbox}{max width=1.\textwidth,center}
\begin{tabular}{l|ccccc|ccccc|ccccc}
\hline
\multirow{2}{*}{Method} & \multicolumn{5}{c|}{in-domain} & \multicolumn{5}{c|}{near-domain} & \multicolumn{5}{c}{out-of-domain} \\
% & BLEU@4 & CIDEr & Fidelity & Adequacy & F1 & BLEU@4 & CIDEr & Fidelity & Adequacy & F1 & BLEU@4 & CIDEr & Fidelity & Adequacy & F1 \\ \hline
 & B@4 & C & P & R & F1 & B@4 & C & P & R & F1 & B@4 & C & P & R & F1 \\ \hline
\textbf{Ours} & \textbf{25.4} & \textbf{87.0} & 58.2 & \textbf{45.6} & \textbf{51.3} & \textbf{22.1} & \textbf{80.1} & 39.9 & \textbf{41.0} & \textbf{40.4} & \textbf{19.7} & \textbf{67.7} & 51.3 & \textbf{30.5} & \textbf{38.3} \\ \hline
Ours w/o $\mathcal{L}_{P}$ & 21.6 & 77.1 & 58.1 & 40.9 & 48 & 19.8 & 75.5 & \textbf{41} & 39.0 & 40.0 & 19.1 & 66.1 & \textbf{51.6} & 19.1 & 66.1 \\
Ours w/o $r_\text{A}$ & 23.0 & 79.3 & \textbf{58.8} & 42.1 & 49.1 & 21.2 & 78.6 & 35.8 & 37.7 & 36.8 & 18.8 & 65.7 & \textbf{51.6} & 27.4 & 35.8 \\
\hline
\end{tabular}
\end{adjustbox}
\vspace{-4mm}
\end{table*}

\vspace{-1.5mm}
\subsection{Ablation studies}
\label{subsec: ablation}
\vspace{-0.5mm}

Following the same evaluation procedures in Sec.~\ref{subsec:metrics}, we discuss the contributions of the uses of paraphrasing model $P$ and association model $A$ in terms of linguistic and semantic level metrics, and present their results in Table~\ref{ablation_table} and~\ref{metric_2}. In addition, to verify the flexibility of our proposed P2C, we replace $P$ and $A$ with different implementations of language models and association models. Then, we evaluate their performance on the nocaps validation set and report the results in Table~\ref{ablation_association_table}.  Detailed ablation analysis of every objective can be further found in Appendix~\ref{appendix:ablation}.

\textbf{Paraphrase model $P$.} As shown in Table~\ref{ablation_table}, the captioning model without $P$ would observe a performance drop in CIDEr for linguistic fluency, but such drops for the metric of SPICE (related to visual content) would be less significant. Similarly, as observed in Table~\ref{metric_2}, removing $P$ would result in the lowest BLEU and CIDEr scores. These results confirm our motivation and model design, since $P$ is utilized to improve caption quality at the linguistics level.

To further verify the use and flexibility of paraphrase model $\mathit{P}$, we replace our paraphrase model $P$ (the~\textit{large} version of BERT) with the~\textit{base} version of BERT. We found out that only a slight performance drop is produced. We conjecture that this is because the two versions of BERT are trained on the same text corpus, which means they share the same word bank. As a result, our model would distill similar linguistic knowledge when using either model to guide the training of our P2C.

\textbf{Semantic-preserving gate $g$.} From Table~\ref{ablation_table}, we see that captioning model would be misled by the wrong guidance produced by $P$, if no validation from $g$ to ensure that the semantics of the original captions is not modified by $P$. This results in a significant performance drop on nocaps.

\textbf{Association model $A$.} As shown in Table~\ref{ablation_table}, the model without the reward calculated by $A$ showed significant drops in captioning metrics of CIDEr and SPICE. However, the performance decrease in SPICE is expected, since $A$ is particularly deployed for preserving visual content in captions. As for CIDEr, its decrease is mainly due to the deterioration of missing visual content in captions. This is also confirmed by Table~\ref{metric_2}, in which mainly the metrics reflecting fidelity and adequacy would observe significant drops for model trained without the association rewards. 

To further verify the use and flexibility of the association model $\mathit{A}$, we replace it with different models that also produce association scores between images and captions. Specifically, we consider VIFIDEL~\cite{madhyastha2019vifidel}, SR-PL~\cite{liu2018show}, and CLIP~\cite{radford2021learning}. As the results shown in Table~\ref{ablation_association_table}, one can see that the use of CLIP as $A$ would achieve the best performance. This is because that, VIFIDEL only associates image-caption data using word embedding of particular object labels, while CLIP assesses such cross-modal data in the instance level, i.e., taking the complete caption of an image into consideration. As for SR-PL, it utilizes the triplet loss to learn the association of images and captions, while CLIP is optimized via the contrastive pre-training, which fully exploits the negative samples in a mini-batch to learn a more compact representation space, allowing it to estimate the association more accurately.

\newcommand{\xmark}{\ding{55}}%
\begin{table}[tp]
\caption{Uses of different paraphrase models $P$ and association models $A$ for NOC. Results on the nocaps validation set are presented. Note that Stages 1 and 2 are the paraphrasing schemes presented in Sec.~\ref{subsec:bert} and Sec.~\ref{subsec:clip}, respectively.}
\label{ablation_association_table}
\setlength{\tabcolsep}{10pt}
\renewcommand{\arraystretch}{1.1}
\begin{adjustbox}{max width=1.0\textwidth, center}
\begin{tabular}{l|c|c|lll|cc}
\hline
\multirow{2}{*}{Method} & \multirow{2}{*}{Paraphrase Model $P$} & \multirow{2}{*}{Association Model $A$} & \multirow{2}{*}{$\mathcal{L}_P$} & \multirow{2}{*}{$r_\text{A}$} & \multirow{2}{*}{$r_\text{rep}$} & \multicolumn{2}{c}{Overall} \\
 &  &  &  &  &  & CIDEr & SPICE \\ \hline
Baseline & \multicolumn{1}{c|}{N/A} & \multicolumn{1}{c|}{N/A} & \xmark & \xmark & \xmark & 81.2 & 12.3 \\
Ours w/ Stage 1 & BERT$_{base}$ & CLIP~\citep{radford2021learning} & $\checkmark$ & \xmark & \xmark & 84.1 & 12.7 \\
Ours w/ Stage 1 & BERT$_{large}$ & CLIP & $\checkmark$ & \xmark & \xmark & 84.2 & 12.7 \\
Ours w/ Stage 2 & BERT$_{large}$ & VIFIDEL~\citep{madhyastha2019vifidel} & $\checkmark$ & $\checkmark$ & $\checkmark$ & 55.3 & 10.7 \\
Ours w/ Stage 2 & BERT$_{large}$ & SR-PL~\citep{liu2018show} & $\checkmark$ & $\checkmark$ & $\checkmark$ & 84.6 & 13.2 \\ \hline
Ours w/ Stage 2 & BERT$_{large}$ & CLIP & $\checkmark$ & $\checkmark$ & $\checkmark$ & \textbf{96.3} & \textbf{14.1} \\ \hline
\end{tabular}
\end{adjustbox}
\end{table}

\textbf{Repetition penalty.}
Recall that, in Sec.~\ref{subsec:clip}, this penalty is to alleviate the association between images and captions with redundant visual information. As seen in Table~\ref{ablation_table}, the model without this penalty observed a significant performance drop in CIDEr scores. We did not see such trends for SPICE. This is because that, repetitive words in captions mainly violate linguistic structures rather then semantic accuracy, and thus the performance related to linguistic fluency would be more sensitive to the deployment of this penalty.
\vspace{0mm}

\begin{figure*}[t]
  \centering
  \includegraphics[page=1,trim={5 258 5 300}, clip, width=0.95\textwidth]{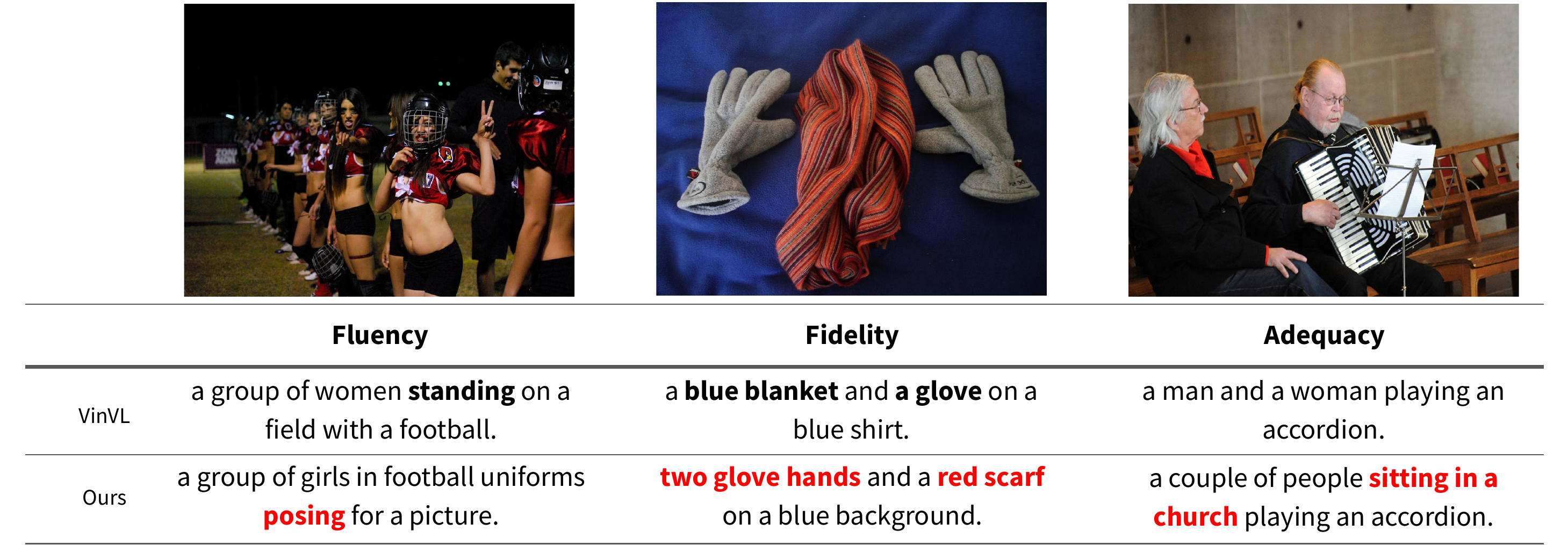}
  \vspace{0mm}
  \caption{Example results and comparisons for image captions produced by VinVL and ours in terms of fluency, fidelity and adequacy. Note that both utilize VIVO for novel object detection.}
  \vspace{0mm}
  \label{fig:qualitative}
\end{figure*}

\vspace{-1mm}
\subsection{Qualitative analysis}
\vspace{-1mm}

For qualitative comparisons, we first conduct human study and ask individuals to evaluate a caption from three perspectives: fluency, fidelity, and adequacy. Due to page limit, the experiment details and the corresponding results are presented in Appendix~\ref{appendix:human_study}. For qualitative analysis, we empirically show captions in Fig.~\ref{fig:qualitative}, which are generated by our model and VinVL, with both pre-trained from VIVO for novel object detection. In this figure, wordings that are less accurate or incorrectly describe the associated visual content are marked in bold. And, our wording improvements are highlighted in red. From this figure, one can see that for fluency, our model generated vivid captions with more proper wordings. Take the image on the left for example, our model particularly described \textit{``posing for a picture''} instead of \textit{``standing on a field''}. As for fidelity, our model is designed to accurately capture the visual content in an image. Specifically, take the second image for example, we correctly described the number of gloves and the novel object \textit{red scarf}, while VinVL failed to do so. As for adequacy, take the bottom-right image for example, our model was able to recover visual details in the image (i.e., \textit{``people playing the accordion''} and \textit{``sitting in a church''}). For more qualitative examples, please refer to Appendix~\ref{appendix:qualitative}.
\vspace{-2mm}

%% file: sections/6_conclusion.tex
\vspace{-2mm}
\section{Conclusion}
\label{sec:conclusion}
\vspace{-2mm}
In this paper, we proposed Paraphrasing-to-Captioning (P2C) for novel object captioning, with particular goals of improving caption fluency, fidelity, and adequacy. In P2C, we advocate the learning of two paraphrasing capabilities for captioning models. The first is language-level paraphrasing, which expands the word bank of a captioning model under the guidance of pre-trained language model, and thus preserves the linguistic fluency during NOC. Secondly, we introduce the self-paraphrasing ability for the captioning model to sufficiently describe visual content of the input image, so that both caption fidelity and adequacy can be achieved. Due to the lack of ground truth captions, image-text association is uniquely exploited for guiding the training process. Empirically, we not only showed that our model achieved SOTA results on benchmark datasets, we also assessed the metrics associated with fluency, fidelity, and adequacy, confirming the effectiveness of our model. Via our ablation studies, we further verified the flexibility of our learning framework by replacing language and cross-modality association modules for paraphrasing and image-text alignment. Finally, since we apply pre-trained language and association models in P2C, their joint optimization would be among our future research directions.

\textbf{Acknowledgement} 
This work is supported in part by the Ministry of Science and Technology of Taiwan under grant NSTC 110-2634-F-002-052. Y-HHT, RS, and L-PM was supported in part by the NSF IIS1763562, NSF Awards \#1750439 \#1722822, National Institutes of Health, IARPA D17PC00340, ONR Grant N000141812861, and Facebook PhD Fellowship. We also thank to National Center for High-performance Computing (NCHC) for providing computational and storage resources.

%% file: sections/7_supp.tex
\newpage
\begin{algorithm}[tp]
\caption{Learning to caption novel objects with linguistic fluency}
\label{alg:alg1}
\SetAlgoLined
 \KwIn{Captioning model $\mathit{C}_\theta(\cdot)$, Paraphrase model $P$, and Association model $A$. }
%  \KwOut{$\theta$}
 \KwData{Captioned image $x_\mathit{l}$, the corresponding GT caption $y_\mathit{l}$, uncaptioned image $x_\mathit{u}$, and lr $\eta_{it}$.}
 \KwOut{Trained Captioning model $\mathit{C}_\theta(\cdot)$.}
 Initialize $\mathit{C}_\theta(\cdot)$; \\
 \For{\textit{it from 1 to num\_iters}}{
    {$\hat{y}_\mathit{l}$ $\gets$ $\mathit{C}_\theta(x_\mathit{l})$},$ $ {$\hat{y}_\mathit{u}^c$ $\gets$ $\mathit{C}_\theta(x_\mathit{u})$}; \\
    Produce $\hat{y}_\mathit{u}^m$ by randomly masking words in the sentence in $\hat{y}_\mathit{u}^c$ (except for nouns); \\
    $\hat{y}_\mathit{u}^p$ $\gets$ P($\hat{y}_\mathit{u}^m$) \\
    $\mathcal{L}_{s2s}$ $\gets$ CrossEntropy($\hat{y}_\mathit{l}, y_\mathit{l}$) \\
    \eIf{ A$(x_\mathit{u}, \hat{y}_\mathit{u}^b)$ $\leq$ A$(x_\mathit{u}, \hat{y}_\mathit{u}^c)$}{
        $\mathcal{L}_{P}$ $\gets$ $0$ 
    }
    {
        
        $\mathcal{L}_{P}$ $\gets$ CrossEntropy($\hat{y}_\mathit{u}^c, \hat{y}_\mathit{u}^p$) 
    }
    $\mathcal{L}$ $\gets$ $\mathcal{L}_{s2s}$ + $\mathcal{L}_{P}$ \\
    Update parameters: $\theta$ $\gets$ \textit{Adam}$(\theta, \eta_{it}, \nabla_{\theta}\mathcal{L})$ 
 }
\end{algorithm}

\section{Remarks on fluency, fidelity, and adequacy}
\label{appendix:connection}
We first discuss how fluency, fidelity, and adequacy can be fundamentally and technically related to language model and association model. For caption fluency, one would expect that the image caption to be linguistically natural and fluent. That is, the NOC model not only requires to capture the occurrence of novel-object vocabularies, the associated collocations such as verbs or modifiers are expected to be properly utilized. Thus, given the context containing novel-object vocabularies as $\tilde{\mathrm{y}}$ and the associated wordings as $\hat{\mathrm{w}}$, we define fluency as the probability of the NOC model which correctly predicts the collocation given the novel image context $p(\hat{\mathrm{w}}|\tilde{\mathrm{y}})$ ($N$ as the number of collocations). In natural language processing, masked language models are widely applied to predict the masked word $\hat{\mathrm{w}}$ given the context $\tilde{\mathrm{y}}$. Thus, the objective of a masked language model would be maximizing the log-likelihood of the masked word $\mathrm{log}(p(\hat{\mathrm{w}}|\tilde{\mathrm{y}}))$ given the context $\tilde{\mathrm{y}}$, which is equivalent to our definition of fluency with the log function explicitly calculated. Therefore, this is the reason why we adopt language model to learn the co-occurrence of novel-object vocabularies and their associated collocations to improve the linguistic fluency.
% \begin{equation}
%     \label{perplexity}
%     \begin{aligned}
%         \sqrt[N]{\frac{1}{p(\hat{\mathrm{y}}_1, \hat{\mathrm{y}}_2, ... \hat{\mathrm{y}}_N|\tilde{\mathrm{y}})}} = (\prod_{i=1}^N p(\hat{\mathrm{y_i}}|\tilde{\mathrm{y}}))^{-\frac{1}{N}}.
%     \end{aligned}
% \end{equation}
% Thus, to achieve high linguistic fluency, we can leverage the language model is expected to produce captions with low perplexity.

We now relate fidelity and adequacy in image captions to cross-modal association. We start by defining the probability of an object appearing in the images as $p(x)$, and the probability of an object mentioned by the captions as $p(y)$. Relevant objects $p(x,y)$ are defined as objects that are both included in the image and described by the associated caption. Since fidelity assesses whether the visual content presented in the produced caption is correct, it can be defined as the fraction of relevant objects among objects in captions $\frac{p(x,y)}{p(y)} = p(x|y)$. On the other hand, adequacy evaluates whether sufficient visual details have been expressed by captions, and it can be defined as the fraction of relevant objects among objects in an image $\frac{p(x,y)}{p(x)} = p(y|x)$. Thus, we can calculate the point-wise mutual information $pmi$ of an image $x$ and its caption $y$ as follows:
\begin{equation}
    \label{mi_eq}
    \begin{aligned}
        pmi(x,y) \equiv log\frac{p(x,y)}{p(x)p(y)} = log\frac{p(x|y)}{p(x)} = log\frac{p(y|x)}{p(y)},
    \end{aligned}
\end{equation}
with mutual information as the expected value of point-wise mutual information. For the task of NOC, both $p(x)$ and $p(y)$ are fixed (i.e., determined by the dataset). Therefore, the above derivation implies that when the mutual information between an image and its captions increases, the resulting fidelity and adequacy would be jointly improved. So that's why we require CLIP to compute the association between images and captions. Since it is trained via the InfoNCE objective, which is a lower bound estimation of mutual information~\citep{oord2018representation}.

\begin{algorithm}[tp]
\caption{Learning novel object captions with fidelity and adequacy}
\label{alg:alg2}
\SetAlgoLined
 \KwIn{Captioning model $\mathit{C}_\theta(\cdot)$ and Association model $A$. }
%  \KwOut{$\theta$}
 \KwData{Captioned image $x_\mathit{l}$, the corresponding GT caption $y_\mathit{l}$, uncaptioned image $x_\mathit{u}$, and lr $\eta_{it}$.}
 \KwOut{Trained Captioning model $\mathit{C}_\theta(\cdot)$.}
 Initialize $\mathit{C}_\theta(\cdot)$; \\
 \For{\textit{it from 1 to num\_iters}}{
    {$\hat{y}_\mathit{l}^s$ $\gets$ $\mathit{C}_\theta(x_\mathit{l})$},$\ $ {$\hat{y}_\mathit{u}^s$ $\gets$ $\mathit{C}_\theta(x_\mathit{u})$} (by sampling);\\
    {$\hat{y}_\mathit{l}^g$ $\gets$ $\mathit{C}_\theta(x_\mathit{l})$},$\ $ {$\hat{y}_\mathit{u}^g$ $\gets$ $\mathit{C}_\theta(x_\mathit{u})$} (by greedy decoding);\\
    Calculate $\mathit{r}_\text{rep}(\hat{y}^s_u)$ and $\mathit{r}_{rep}(\hat{y}^g_u)$ by (\ref{rep_reward})\\
    $\mathit{r}(\hat{y}_l)$ $\gets$ $\mathit{r}_\text{CIDEr}(\hat{y}_l, y_l)$ + $\mathit{r}_\text{A}({x}_l, \hat{y}_l)$ \\
    $\mathit{r}(\hat{y}_u)$ $\gets$ $\mathit{r}_\text{A}({x}_u, \hat{y}_u)$ + $\mathit{r}_\text{rep}(\hat{y}_u)$ \\
    Calculate the gradient $\nabla_{\theta}\mathcal{L}_{RL}(\theta)$ $\gets$ $-(\mathit{r}(\hat{y}^s_d)-\mathit{r}(\hat{y}^g_d))\nabla_{\theta}\log p_{\theta}(\hat{y}^s_d)$, $\mathit{d} \in \{l, u\}$ \\
    Update parameters: $\theta$ $\gets$ \textit{Adam}$(\theta,\ \eta_{it},\ \nabla_{\theta}\mathcal{L}_{RL})$ 
 }
\end{algorithm}

\begin{table*}[tp]
\small
\caption{Ablation studies on nocaps validation set.}
\label{ablation_incremental}
\renewcommand{\arraystretch}{1.5}
\begin{adjustbox}{max width=1.0\textwidth,center}
\begin{tabular}{cccccrcrc}
\hline
\multicolumn{1}{l|}{\multirow{2}{*}{Method}} & \multicolumn{2}{c|}{in-domain} & \multicolumn{2}{c|}{near-domain} & \multicolumn{2}{c|}{out-of-domain} & \multicolumn{2}{c}{overall} \\ 
\multicolumn{1}{c|}{} & CIDEr & \multicolumn{1}{c|}{SPICE} & CIDEr & \multicolumn{1}{c|}{SPICE} & \multicolumn{1}{c}{CIDEr} & \multicolumn{1}{c|}{SPICE} & \multicolumn{1}{c}{CIDEr} & SPICE \\ \hline
\multicolumn{1}{l|}{Baseline (Only w/ $L_{s2s}$)} & 89.07 & \multicolumn{1}{c|}{13.29} & 83.29 & \multicolumn{1}{c|}{12.61} & 68.77 & \multicolumn{1}{c|}{10.59} & 81.17 & 12.32 \\ 
\multicolumn{1}{l|}{+ $L_{P}$} & 92.46 & \multicolumn{1}{c|}{13.40} & 85.79 & \multicolumn{1}{c|}{12.92} & 73.21 & \multicolumn{1}{c|}{11.40} & 84.20 & 12.69 \\ 
\multicolumn{1}{l|}{+ $r_\text{CIDEr}$} & 101.19 & \multicolumn{1}{c|}{13.84} & 95.38 & \multicolumn{1}{c|}{13.44} & 83.24 & \multicolumn{1}{c|}{12.06} & 93.75 & 13.23\\ 
\multicolumn{1}{l|}{+ $r_\text{A}$} & 96.73 & \multicolumn{1}{c|}{\textbf{14.83}} & 89.64 & \multicolumn{1}{c|}{14.12} & 81.87 & \multicolumn{1}{c|}{12.38} & 89.08 & 13.88 \\
\multicolumn{1}{l|}{+ $r_\text{rep}$ (Ours)} & \textbf{102.77} & \multicolumn{1}{c|}{\textbf{14.83}} & \textbf{97.90} & \multicolumn{1}{c|}{\textbf{14.40}} & \textbf{86.33} & \multicolumn{1}{c|}{\textbf{12.54}} & \textbf{96.25} & \textbf{14.10} \\ \hline
\end{tabular}
\end{adjustbox}
\end{table*}

\section{Implementation details}
\label{appendix:implementation}
Following~\citet{hu2020vivo,li2020oscar, zhang2021vinvl}, we consider a BERT-base~\citep{devlin2018bert} architecture for our captioning model. Given an image, the captioning model jointly takes the image region features and the predicted detection tags to generate the associated caption. We use the same region features as VinVL~\citep{zhang2021vinvl}, which are released on their project page. Since the object detection model Omni-detection used in previous works~\citep{hu2020vivo, zhang2021vinvl} is not available, we replace it with a publicly available model of TSD~\citep{song2020revisiting} to generate the object detection tag.

\textbf{Reproducing our method.} We perform VIVO~\citep{hu2020vivo} pre-training for 100 epochs with a batch size of 1024 and a learning rate of $5 \times 10^{-5}$, which are exactly the same as the parameters stated in the VIVO paper. After that, we propose to train our model following the training process described in Algorithm~\ref{alg:alg1} to learn to caption novel objects with linguistic fluency. We train our model for 20 epochs with an effective batch size of 512 (256 caption-labeled images and 256 uncaptioned images) and a learning rate of $1.5 \times 10^{-5}$. Then, to learn novel object captions with fidelity and adequacy, we train our model as decsribed in Algorithm~\ref{alg:alg2}. Specifically, we train our model for 4 epochs with an effective batch size of 128 (64 caption-labeled images and 64 uncaptioned images) and a learning rate of $2.5 \times 10^{-6}$. We use 8 V100 GPUs to perform the above training algorithms. Codes can be found in the supplementary materials.

\textbf{Reproducing baseline methods.} For VinVL~\citep{zhang2021vinvl}, we leverage the released model on their project page and directly inference on the nocaps dataset. However, for VinVL+VIVO~\citep{zhang2021vinvl}, since the pre-trained model is not publicily available, we reproduce this method using the image region features and object detection tags generated by models mentioned in the beginning of this section to train this model. Specifically, the model is trained for 160K iterations (about 100 epochs) with a batch size of 1024 and a learning rate of $5 \times 10^{-5}$, and fine-tuned for 30 epochs with a batch size of 256 and a learning rate of $5 \times 10^{-5}$ using the cross-entropy loss. Last, we perform the SCST optimization~\citep{rennie2017self} with a
learning rate of $2 \times 10^{-6}$ for 5 epochs to obtain the final model. The numbers reported in Table~\ref{ablation_incremental} are derived using this version of model.

\section{Additional experiments}

\subsection{Detailed ablation analysis}
\label{appendix:ablation}
Table~\ref{ablation_incremental} lists the performances and compares contributions of the imposed objectives in our P2C. The baseline model in Table~\ref{ablation_incremental} is only trained on the COCO Caption dataset using the sequence-to-sequence objective. To confirm our introduction of exploiting BERT to learn the associated wordings of novel images, we apply $L_{P}$ to the baseline model, and report the results in the second row of Table~\ref{ablation_incremental}. The CIDEr scores improve significantly after adopting reinforce algorithm~\citep{williams1992simple} and using CIDEr scores of the generated captions as reward, and the results are in the third row. One can see that the SPICE scores largely increase but the CIDEr scores slightly decrease after the deployment of the association model $A$. We hypothesize that the captioning model properly captures the visual content in images, but it describes the scene with poor linguistic fluency. As the discussion in Sec.~\ref{subsec:clip}, we attribute the performance drop to the degenerate solution of increasing the association between the captions and the corresponding images. Note that we further consider the repetition penalty to regularize the captioning model. The results are shown in the last row of Table~\ref{ablation_incremental}. One can see that this regularization slightly improves the SPICE scores but significantly increase the CIDEr scores. By comparing the performances listed in Table~\ref{ablation_incremental}, we see that the full version of our P2C achieved the best performance in terms of CIDEr and SPICE. Thus, the design of our P2C can be successfully verified.

\subsection{Experiments on the COCO Caption dataset}
\label{appendix:cocoexp}
\begin{table}[tp]
\caption{Image captioning evaluation results on COCO “Karpathy” test split. Note that B@4 stands for BLEU@4, M for METEOR, R for ROUGE-L, C for CIDEr, and S for SPICE.}
\label{coco_table}
\renewcommand{\arraystretch}{1.25}
\begin{adjustbox}{max width=1.0\textwidth,center}
\begin{tabular}{l|r|r|r|r|r}
\hline
 & \multicolumn{1}{c|}{B@4} & \multicolumn{1}{c|}{M} & \multicolumn{1}{c|}{R} & \multicolumn{1}{c|}{C} & \multicolumn{1}{c}{S} \\ \hline
VinVL & 39.8 & 29.9 & 59.6 & 134.6 & 23.9 \\
VinVL+VIVO & 39.7 & 29.9 & 59.6 & 134.5 & 23.8 \\ \hline
\textbf{Ours} & \textbf{40.0} & \textbf{30.4} & \textbf{60.2} & \textbf{137.3} & \textbf{24.5} \\ \hline
\end{tabular}
\end{adjustbox}
\end{table}

\begin{table*}[tp]
\small
\caption{Ablation studies of the joint-training model on nocaps validation set.}
\label{ablation_xd}
\renewcommand{\arraystretch}{1.15}
\begin{adjustbox}{max width=1.0\textwidth,center}
\begin{tabular}{cccccrcrc}
\hline
\multicolumn{1}{l|}{\multirow{2}{*}{Method}} & \multicolumn{2}{c|}{in-domain} & \multicolumn{2}{c|}{near-domain} & \multicolumn{2}{c|}{out-of-domain} & \multicolumn{2}{c}{overall} \\ 
\multicolumn{1}{c|}{} & CIDEr & \multicolumn{1}{c|}{SPICE} & CIDEr & \multicolumn{1}{c|}{SPICE} & \multicolumn{1}{c}{CIDEr} & \multicolumn{1}{c|}{SPICE} & \multicolumn{1}{c}{CIDEr} & SPICE \\ \hline
\multicolumn{1}{l|}{Baseline (Only w/ $L_{s2s}$)} & 96.1 & \multicolumn{1}{c|}{13.71} & 90.35 & \multicolumn{1}{c|}{13.41} & 79.96 & \multicolumn{1}{c|}{11.77} & 89.07 & 13.13 \\
\multicolumn{1}{l|}{+ $L_{P}$} & 99.44 & \multicolumn{1}{c|}{13.91} & 91.13 & \multicolumn{1}{c|}{13.53} & 81.11 & \multicolumn{1}{c|}{11.82} & 90.29 & 13.25 \\
\multicolumn{1}{l|}{+ $r_\text{CIDEr}$} & 109.14 & \multicolumn{1}{c|}{14.52} & 100.66 & \multicolumn{1}{c|}{14.08} & 88.61 & \multicolumn{1}{c|}{12.69} & 99.43 & 13.87 \\
\multicolumn{1}{l|}{+ $r_\text{A}$} & 103.81 & \multicolumn{1}{c|}{\textbf{15.99}} & 98.91 & \multicolumn{1}{c|}{\textbf{15.32}} & 93.17 & \multicolumn{1}{c|}{\textbf{13.67}} & 98.45 & \textbf{15.09} \\
\multicolumn{1}{l|}{+ $r_\text{rep}$ (Ours)} & \textbf{110.56} & \multicolumn{1}{c|}{15.23} & \textbf{105.16} & \multicolumn{1}{c|}{14.81} & \textbf{96.22} & \multicolumn{1}{c|}{13.19} & \textbf{104.12} & 14.55 \\ \hline
\end{tabular}
\end{adjustbox}
\end{table*}

% \begin{table}[tp]
% \caption{Quantitative results on the nocaps (XD) test set.}
% \label{xd_table}
% \renewcommand{\arraystretch}{1.2}
% \begin{adjustbox}{max width=1.0\textwidth,center}
% \begin{tabular}{ccc}
% \hline
% \multicolumn{1}{l|}{\multirow{2}{*}{Method}} & \multicolumn{2}{c}{overall} \\ 
% \multicolumn{1}{c|}{} & \multicolumn{1}{c}{CIDEr} & SPICE \\ \hline
% \multicolumn{1}{l|}{UpDown~\citep{agrawal2019nocaps}} &  73.09 & 11.20 \\ 
% \multicolumn{1}{l|}{$\text{SimVLM}_\text{base}$~\citep{wang2021simvlm}} & 94.80 & 13.10 \\ 
% \multicolumn{1}{l|}{VIVO~\citep{hu2020vivo}} & 100.12 & 14.04 \\ \hline
% \multicolumn{1}{l|}{Ours} &  96.25 & 14.10 \\ 
% \multicolumn{1}{l|}{Ours (+CC)} &  \textbf{102.39} & \textbf{14.71} \\ \hline
% \end{tabular}
% \end{adjustbox}
% \end{table}

To validate that our method generalize well on the task of describing the seen objects, we conduct experiments on the COCO Caption test set and report the numbers in Table~\ref{coco_table}. The training data for VinVL~\citep{zhang2021vinvl} is image caption pairs from the COCO~\citep{lin2014microsoft} dataset. While for VinVL + VIVO and our method, we additionally leverage the uncaptioned image from the Open Images~\citep{kuznetsova2020open} dataset as extra data. One can see that our method outperforms the other competitive approaches on different metrics which verifies the effectiveness of our approach.

\subsection{Experiments on the nocaps (XD) benchmark}
\label{appendix:nocapsxd}
% To investigate the limits of performance on nocaps without any restraints on the training datasets, we conduct experiments on the nocaps (XD) benchmark to verify the effectiveness of our method when more image-caption pairs are considered. Specifically, we additionally consider Conceptual Captions (CC)~\citep{sharma2018conceptual} as labeled training samples $X_l$ in our learning framework and perform joint-training to see if extra image-caption pairs benefit the model on novel object captioning. The results are shown in Table~\ref{xd_table}. Note that both VIVO, SimVLM and our method adopt BERT-based architecture as the backbones for captioning. As for the training set, since VIVO~\citep{hu2020vivo} did not specify the dataset details for their evaluation for the nocaps (XD), we compared our method to SimVLM, which applied a much larger web-scale dataset (1.8B image-text pairs) than the CC dataset (3.1M pairs). Yet, our method still performs favorably against SOTAs on the nocaps (XD) protocol and benchmark, verifying the effectiveness of our method even if more image-caption pairs are considered.
Recall that in Sec. 4.2, we quantitatively show that our method surpasses current state-of-the-art large-scale methods even if we use a smaller training corpus. In the subsection, we would like to investigate whether the improvement from our module design is still significant when we scale up training data.

To quantitatively show that the performance gain in Table~\ref{nocapsxd_table} is not simply contributed by the additional data we considered, we ablate our model on the nocaps validation set and show the results in Table~\ref{ablation_xd}. We observed a similar performance trend as we reported in Table~\ref{ablation_incremental}, where $L_{P}$ slightly improves the CIDEr scores, and $r_\text{A}$ significantly boost SPICE but slightly deteriorates the CIDEr scores. One can see that the regularization $r_\text{rep}$ slightly improves the SPICE scores but significantly increase the CIDEr. By comparing the performances listed in Table~\ref{ablation_incremental} and Table~\ref{ablation_xd}, we see that our design of using a paraphrase model $P$ to enhance fluency (in terms of CIDEr) and the uses of the association model $A$ to encourage captions with sufficient fidelity and adequacy (in terms of SPICE) still function properly when more diverse image-caption pairs are considered, verifying the design of our P2C.

% \subsection{Analysis on the association module $\mathit{A}$}
% \label{appendix:association_module}
% To further verify the model selection of our association module $\mathit{A}$, we compare the quantitative performances of our P2C with different association modules. Specifically, we consider DISC~\cite{luo2018discriminability} and VIFIDEL~\cite{madhyastha2019vifidel} as possible candidates of the association module $\mathit{A}$ and replace CLIP with these models for comparisons. As the results shown in Table~\ref{ablation_association_table}, one can see that the use of VIFIDEL was not able to achieve comparable results as that of CLIP, and the use of DISC. That is because VIFIDEL only associates image-caption data using word embeddings of particular object labels, while CLIP assesses such cross-modal data in the instance level, i.e., taking the complete caption of an image into consideration.

\begin{table}[t]
\caption{Human study on the nocaps validation set.}
\label{human_study}
\renewcommand{\arraystretch}{1.2}
\begin{adjustbox}{max width=1.0\textwidth,center}
\setlength{\tabcolsep}{0.6mm}{
\begin{tabular}{ccccc}
\hline
\multicolumn{1}{l|}{Method} & \multicolumn{1}{c|}{M1} & \multicolumn{1}{c|}{M2} & \multicolumn{1}{c|}{M3} & M4 \\
\multicolumn{1}{l|}{} & \multicolumn{1}{c|}{(Turing Test)} & \multicolumn{1}{c|}{(Fluency)} & \multicolumn{1}{c|}{(Fidelity)} & (Adequacy) \\
\hline
\multicolumn{1}{l|}{VinVL} &  \multicolumn{1}{c|}{\multirow{2}{*}{0.25}} & \multicolumn{1}{c|}{\multirow{2}{*}{3.99}} & \multicolumn{1}{c|}{\multirow{2}{*}{3.70}} & \multirow{2}{*}{3.46} \\
\multicolumn{1}{l|}{\ \ +VIVO} & \multicolumn{1}{c|}{} & \multicolumn{1}{c|}{} & \multicolumn{1}{c|}{} &  \\
\multicolumn{1}{l|}{Ours} & \multicolumn{1}{c|}{0.43} & \multicolumn{1}{c|}{4.06} & \multicolumn{1}{c|}{4.33} & \textbf{4.24} \\ 
\multicolumn{1}{l|}{Human} & \multicolumn{1}{c|}{\textbf{0.53}} & \multicolumn{1}{c|}{\textbf{4.09}} & \multicolumn{1}{c|}{\textbf{4.44}} & 4.18 \\ \hline
\end{tabular}}
\end{adjustbox}
\end{table}

\subsection{Human study}
\label{appendix:human_study}
To conduct human study, we randomly picked 60 images from the nocaps validation set, and compared the captions generated by our method to those generated by the SOTA of VinVL+VIVO~\citep{zhang2021vinvl}, and the human-annotated captions provided by the nocaps dataset. Following the evaluation protocols used in the COCO Captioning Challenge 2015~\citep{lin2014microsoft}, we designed 4 different metrics and asked individuals to evaluate captions from these aspects. The following are the four metrics we used in the experiment:
M1: Is the caption generated by human (0: machine, 1: human)? (Percentage of captions that pass the Turing Test.)
M2: Rate the correctness of the captions on a scale 1-5 (incorrect-correct): Whether the described objects or activities are correct.
M3: Rate the amount of detail of the captions on a scale 1-5 (lack of details - very detailed): Whether the caption has detailed all the objects and their attributes.
M4: Rate the fluency of the captions on a scale 1-5 (lack of fluency-very fluent): Whether the caption use phrases/words that human generally would use to describe the scene, i.e., the caption is linguistically natural and fluent.

Specifically, M2, M3, M4 correspond to the fidelity, adequacy, and fluency, respectively, which are the particular objectives desired to be achieved. We asked 24 people two answer 6 different questionnaires, and each questionnaire contains 10 captions from each method (i.e., ours, sota, and human caption presented in a random order). We report the results in Table~\ref{human_study}. We see that our method surpassed the SOTA by clear margins, while our performances were comparable to those the human ones across different metrics. This further supports the design of our model for NOC with sufficient fluency, fidelity, and adequacy.

\subsection{More qualitative results}
\label{appendix:qualitative}
\begin{figure*}[t]
  \centering
  \includegraphics[page=2,trim={5 130 5 10}, clip, width=\textwidth]{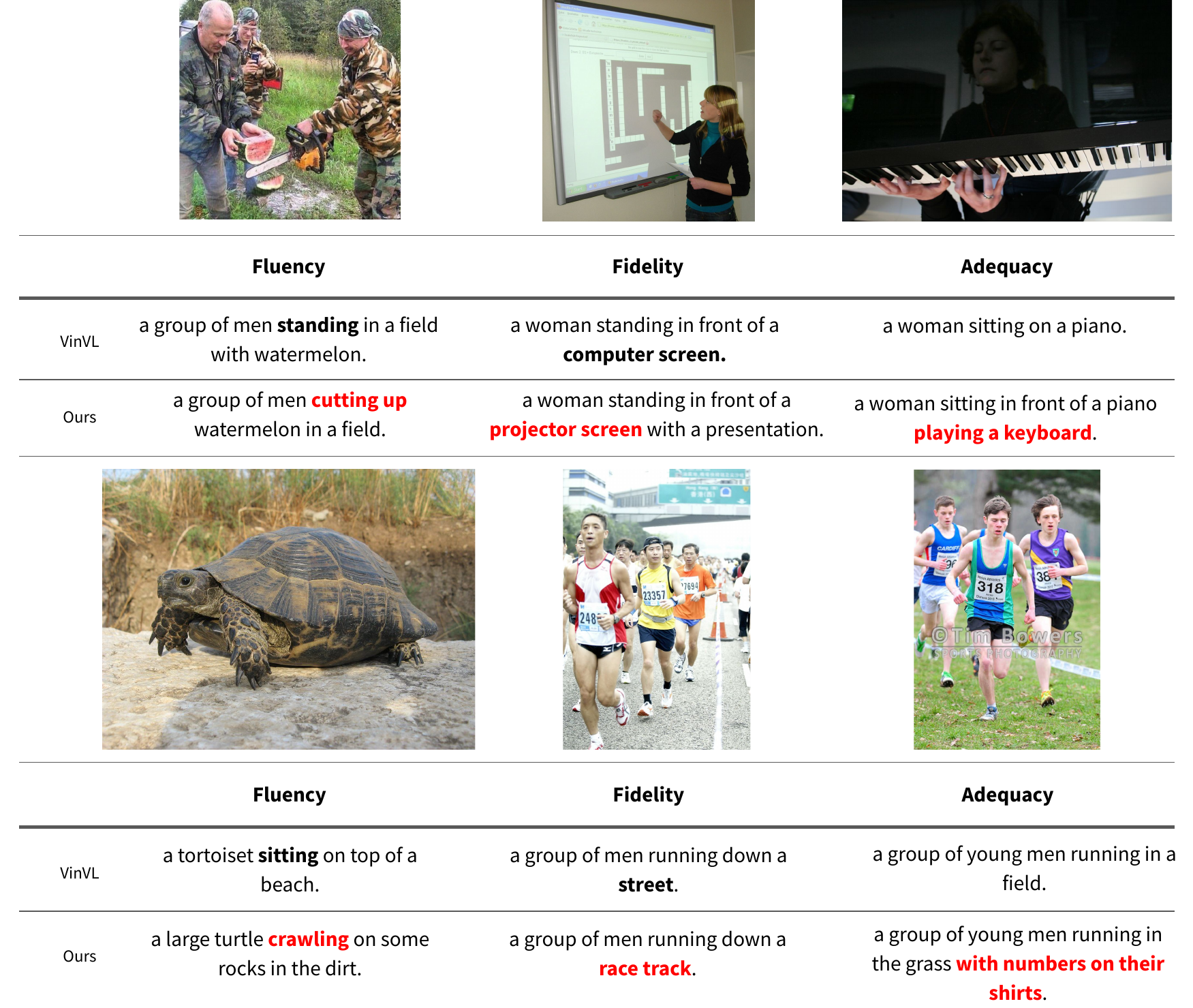}
  \vspace{-4mm}
  \caption{Example results and comparisons for image captions produced by VinVL and ours in terms of fluency, fidelity and adequacy. Note that both utilize VIVO for novel object detection.}
  \vspace{-4mm}
  \label{appendix:fig:qualitative_0}
\end{figure*}

\begin{figure*}[t]
  \centering
  \includegraphics[page=3,trim={5 135 5 10}, clip, width=\textwidth]{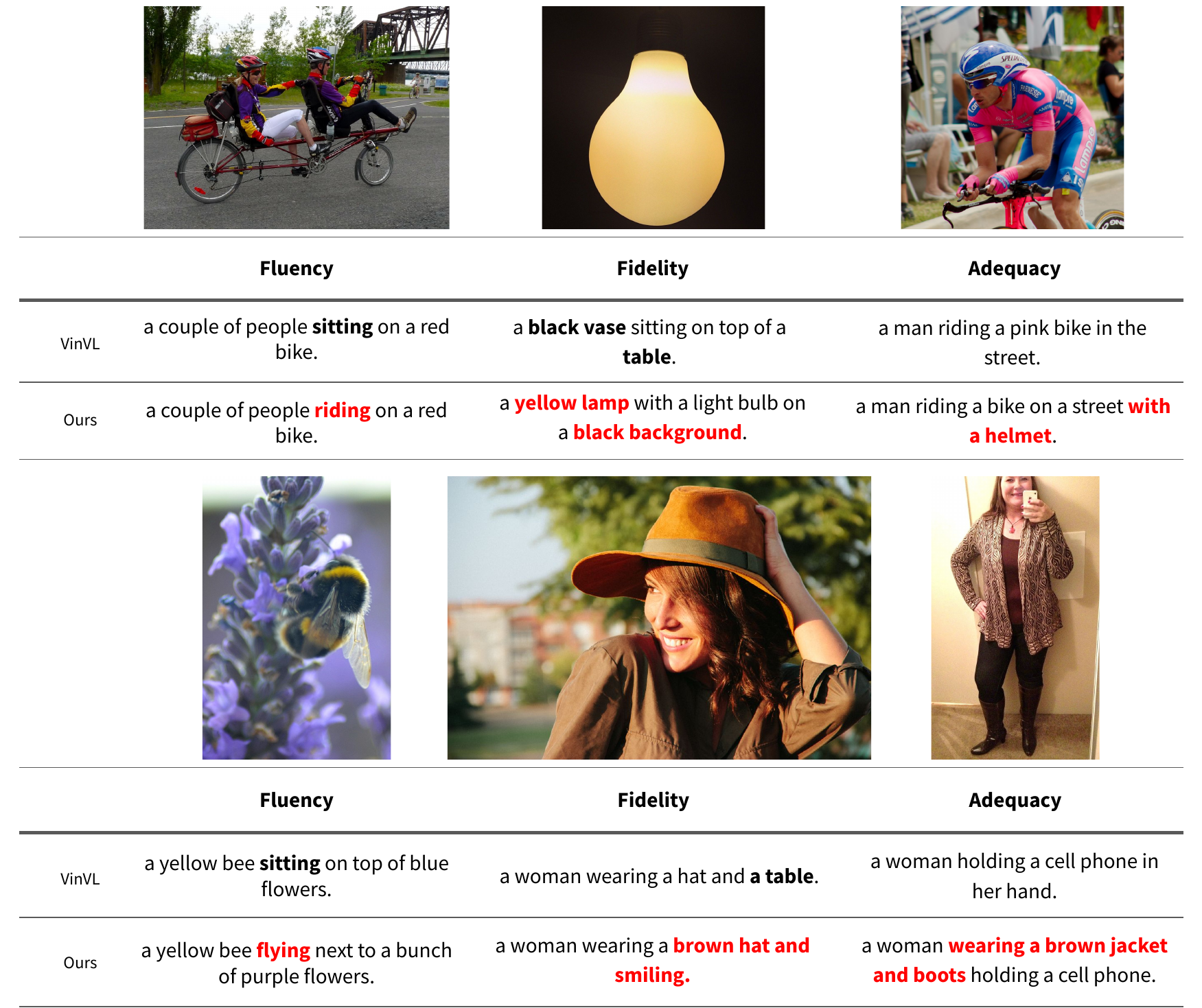}
  \vspace{-4mm}
  \caption{Example results and comparisons for image captions produced by VinVL and ours in terms of fluency, fidelity and adequacy. Note that both utilize VIVO for novel object detection.}
  \vspace{-4mm}
  \label{appendix:fig:qualitative_1}
\end{figure*}

\textbf{Qualitative comparison on fluency, fidelity and adequacy.} In this part, we provide more qualitative results on the nocaps validation/test set, and the results are shown in Fig.~\ref{appendix:fig:qualitative_0} and~\ref{appendix:fig:qualitative_1}. Note that wordings that are less accurate or incorrectly describe the associated visual content are marked in bold. And, our wording improvements are highlighted in red. Take results in the bottom row of Fig.~\ref{appendix:fig:qualitative_0} for example. For the column of fluency, our model particularly described the turtle being \textit{``crawling on some rocks"} instead of \textit{``sitting on the top of a beach"}. For fidelity, our model predicted the background preferably as \textit{``race track''} instead of \textit{``street"} from the prediction of VinVL model. As for the column of adequacy, though both captions described a young men running, our model successfully captures more details in the image (i.e., \textit{``there are number on their shirts"}). For more qualitative results, please refer to Fig.~\ref{appendix:fig:qualitative_1}.

\begin{figure*}[t]
  \centering
  \includegraphics[page=4,trim={5 180 5 300}, clip, width=\textwidth]{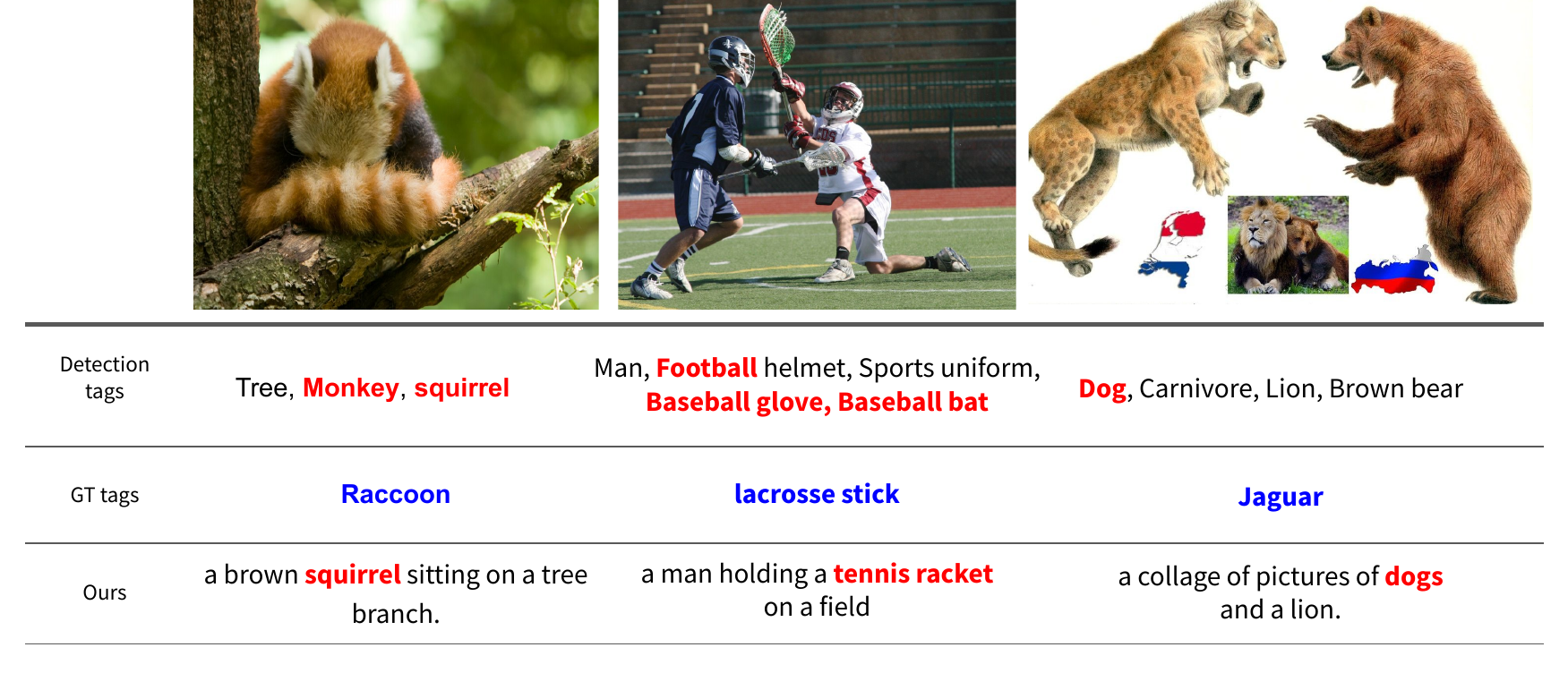}
  \vspace{-4mm}
  \caption{False captions misled by the wrong object detection tags.}
  \vspace{-4mm}
  \label{appendix:fig:qualitative_2}
\end{figure*}

\textbf{Qualitative results of some failure cases.} In this part, we demonstrate some failure cases of our P2C model. We empirically observe that the failure cases mainly come from the wrong/missing detection tags predicted by the pre-trained object detectors. To be more specific, the captioning model largely relies on the detection tags as clues to correctly describe novel objects. Take the result in the left-side of Fig.~\ref{appendix:fig:qualitative_2} for example, the detection model falsely recognizes the raccoon as a squirrel, and this detection result consequently damages the caption prediction. Therefore, how to jointly improve the detection model and captioning model is still a open question, and we leave this problem for future research. For more failure cases, please refer to Fig.~\ref{appendix:fig:qualitative_2}.

%% file: main.bbl
\begin{thebibliography}{45}
\providecommand{\natexlab}[1]{#1}
\providecommand{\url}[1]{\texttt{#1}}
\expandafter\ifx\csname urlstyle\endcsname\relax
  \providecommand{\doi}[1]{doi: #1}\else
  \providecommand{\doi}{doi: \begingroup \urlstyle{rm}\Url}\fi

\bibitem[Agrawal et~al.(2019)Agrawal, Desai, Wang, Chen, Jain, Johnson, Batra,
  Parikh, Lee, and Anderson]{agrawal2019nocaps}
Harsh Agrawal, Karan Desai, Yufei Wang, Xinlei Chen, Rishabh Jain, Mark
  Johnson, Dhruv Batra, Devi Parikh, Stefan Lee, and Peter Anderson.
\newblock nocaps: novel object captioning at scale.
\newblock In \emph{Proceedings of the IEEE/CVF International Conference on
  Computer Vision}, pages 8948--8957, 2019.

\bibitem[Chen et~al.(2015)Chen, Fang, Lin, Vedantam, Gupta, Doll{\'a}r, and
  Zitnick]{chen2015microsoft}
Xinlei Chen, Hao Fang, Tsung-Yi Lin, Ramakrishna Vedantam, Saurabh Gupta, Piotr
  Doll{\'a}r, and C~Lawrence Zitnick.
\newblock Microsoft coco captions: Data collection and evaluation server.
\newblock \emph{arXiv preprint arXiv:1504.00325}, 2015.

\bibitem[Young et~al.(2014)Young, Lai, Hodosh, and Hockenmaier]{young2014image}
Peter Young, Alice Lai, Micah Hodosh, and Julia Hockenmaier.
\newblock From image descriptions to visual denotations: New similarity metrics
  for semantic inference over event descriptions.
\newblock \emph{Transactions of the Association for Computational Linguistics},
  2:\penalty0 67--78, 2014.

\bibitem[Huang et~al.(2019)Huang, Wang, Chen, and Wei]{huang2019attention}
Lun Huang, Wenmin Wang, Jie Chen, and Xiao-Yong Wei.
\newblock Attention on attention for image captioning.
\newblock In \emph{Proceedings of the IEEE/CVF International Conference on
  Computer Vision}, pages 4634--4643, 2019.

\bibitem[Wang et~al.(2019)Wang, Chen, and Hu]{wang2019hierarchical}
Weixuan Wang, Zhihong Chen, and Haifeng Hu.
\newblock Hierarchical attention network for image captioning.
\newblock In \emph{Proceedings of the AAAI Conference on Artificial
  Intelligence}, volume~33, pages 8957--8964, 2019.

\bibitem[Guo et~al.(2020)Guo, Liu, Zhu, Yao, Lu, and Lu]{guo2020normalized}
Longteng Guo, Jing Liu, Xinxin Zhu, Peng Yao, Shichen Lu, and Hanqing Lu.
\newblock Normalized and geometry-aware self-attention network for image
  captioning.
\newblock In \emph{Proceedings of the IEEE/CVF Conference on Computer Vision
  and Pattern Recognition}, pages 10327--10336, 2020.

\bibitem[Cornia et~al.(2020)Cornia, Stefanini, Baraldi, and
  Cucchiara]{cornia2020meshed}
Marcella Cornia, Matteo Stefanini, Lorenzo Baraldi, and Rita Cucchiara.
\newblock Meshed-memory transformer for image captioning.
\newblock In \emph{Proceedings of the IEEE/CVF Conference on Computer Vision
  and Pattern Recognition}, pages 10578--10587, 2020.

\bibitem[Zhou et~al.(2020)Zhou, Palangi, Zhang, Hu, Corso, and
  Gao]{zhou2020unified}
Luowei Zhou, Hamid Palangi, Lei Zhang, Houdong Hu, Jason Corso, and Jianfeng
  Gao.
\newblock Unified vision-language pre-training for image captioning and vqa.
\newblock In \emph{Proceedings of the AAAI Conference on Artificial
  Intelligence}, volume~34, pages 13041--13049, 2020.

\bibitem[Gu et~al.(2019)Gu, Joty, Cai, Zhao, Yang, and Wang]{gu2019unpaired}
Jiuxiang Gu, Shafiq Joty, Jianfei Cai, Handong Zhao, Xu~Yang, and Gang Wang.
\newblock Unpaired image captioning via scene graph alignments.
\newblock In \emph{Proceedings of the IEEE/CVF International Conference on
  Computer Vision}, pages 10323--10332, 2019.

\bibitem[Feng et~al.(2019)Feng, Ma, Liu, and Luo]{feng2019unsupervised}
Yang Feng, Lin Ma, Wei Liu, and Jiebo Luo.
\newblock Unsupervised image captioning.
\newblock In \emph{Proceedings of the IEEE/CVF Conference on Computer Vision
  and Pattern Recognition}, pages 4125--4134, 2019.

\bibitem[Lu et~al.(2018)Lu, Yang, Batra, and Parikh]{lu2018neural}
Jiasen Lu, Jianwei Yang, Dhruv Batra, and Devi Parikh.
\newblock Neural baby talk.
\newblock In \emph{Proceedings of the IEEE conference on computer vision and
  pattern recognition}, pages 7219--7228, 2018.

\bibitem[Wu et~al.(2018)Wu, Zhu, Jiang, and Yang]{wu2018decoupled}
Yu~Wu, Linchao Zhu, Lu~Jiang, and Yi~Yang.
\newblock Decoupled novel object captioner.
\newblock In \emph{Proceedings of the 26th ACM international conference on
  Multimedia}, pages 1029--1037, 2018.

\bibitem[Hu et~al.(2020)Hu, Yin, Lin, Wang, Zhang, Gao, and Liu]{hu2020vivo}
Xiaowei Hu, Xi~Yin, Kevin Lin, Lijuan Wang, Lei Zhang, Jianfeng Gao, and
  Zicheng Liu.
\newblock Vivo: Surpassing human performance in novel object captioning with
  visual vocabulary pre-training.
\newblock \emph{arXiv e-prints}, pages arXiv--2009, 2020.

\bibitem[Rennie et~al.(2017)Rennie, Marcheret, Mroueh, Ross, and
  Goel]{rennie2017self}
Steven~J Rennie, Etienne Marcheret, Youssef Mroueh, Jerret Ross, and Vaibhava
  Goel.
\newblock Self-critical sequence training for image captioning.
\newblock In \emph{Proceedings of the IEEE conference on computer vision and
  pattern recognition}, pages 7008--7024, 2017.

\bibitem[Li et~al.(2019)Li, Chen, and Liu]{li2019meta}
Nannan Li, Zhenzhong Chen, and Shan Liu.
\newblock Meta learning for image captioning.
\newblock In \emph{Proceedings of the AAAI Conference on Artificial
  Intelligence}, volume~33, pages 8626--8633, 2019.

\bibitem[Yang et~al.(2020)Yang, Zhang, Jin, Liu, Wu, Tan, Xie, Wang, and
  Wang]{yang2020fashion}
Xuewen Yang, Heming Zhang, Di~Jin, Yingru Liu, Chi-Hao Wu, Jianchao Tan,
  Dongliang Xie, Jue Wang, and Xin Wang.
\newblock Fashion captioning: Towards generating accurate descriptions with
  semantic rewards.
\newblock In \emph{Computer Vision--ECCV 2020: 16th European Conference,
  Glasgow, UK, August 23--28, 2020, Proceedings, Part XIII 16}, pages 1--17.
  Springer, 2020.

\bibitem[Liu et~al.(2018)Liu, Li, Shao, Chen, and Wang]{liu2018show}
Xihui Liu, Hongsheng Li, Jing Shao, Dapeng Chen, and Xiaogang Wang.
\newblock Show, tell and discriminate: Image captioning by self-retrieval with
  partially labeled data.
\newblock In \emph{Proceedings of the European Conference on Computer Vision
  (ECCV)}, pages 338--354, 2018.

\bibitem[Kim et~al.(2019)Kim, Choi, Oh, and Kweon]{kim2019image}
Dong-Jin Kim, Jinsoo Choi, Tae-Hyun Oh, and In~So Kweon.
\newblock Image captioning with very scarce supervised data: Adversarial
  semi-supervised learning approach.
\newblock \emph{arXiv preprint arXiv:1909.02201}, 2019.

\bibitem[Anderson et~al.(2016{\natexlab{a}})Anderson, Fernando, Johnson, and
  Gould]{anderson2016guided}
Peter Anderson, Basura Fernando, Mark Johnson, and Stephen Gould.
\newblock Guided open vocabulary image captioning with constrained beam search.
\newblock \emph{arXiv preprint arXiv:1612.00576}, 2016{\natexlab{a}}.

\bibitem[Anderson et~al.(2018)Anderson, Gould, and
  Johnson]{anderson2018partially}
Peter Anderson, Stephen Gould, and Mark Johnson.
\newblock Partially-supervised image captioning.
\newblock \emph{arXiv preprint arXiv:1806.06004}, 2018.

\bibitem[Hendricks et~al.(2016)Hendricks, Venugopalan, Rohrbach, Mooney,
  Saenko, and Darrell]{hendricks2016deep}
Lisa~Anne Hendricks, Subhashini Venugopalan, Marcus Rohrbach, Raymond Mooney,
  Kate Saenko, and Trevor Darrell.
\newblock Deep compositional captioning: Describing novel object categories
  without paired training data.
\newblock In \emph{Proceedings of the IEEE conference on computer vision and
  pattern recognition}, pages 1--10, 2016.

\bibitem[Yao et~al.(2017)Yao, Pan, Li, and Mei]{yao2017incorporating}
Ting Yao, Yingwei Pan, Yehao Li, and Tao Mei.
\newblock Incorporating copying mechanism in image captioning for learning
  novel objects.
\newblock In \emph{Proceedings of the IEEE conference on computer vision and
  pattern recognition}, pages 6580--6588, 2017.

\bibitem[Vo et~al.(2022)Vo, Chen, Sugimoto, and Nakayama]{vo2022noc}
Duc~Minh Vo, Hong Chen, Akihiro Sugimoto, and Hideki Nakayama.
\newblock Noc-rek: Novel object captioning with retrieved vocabulary from
  external knowledge.
\newblock \emph{arXiv preprint arXiv:2203.14499}, 2022.

\bibitem[Wang et~al.(2021{\natexlab{a}})Wang, Yao, Wang, Wu, and
  Chen]{wang2021faier}
Sijin Wang, Ziwei Yao, Ruiping Wang, Zhongqin Wu, and Xilin Chen.
\newblock Faier: Fidelity and adequacy ensured image caption evaluation.
\newblock In \emph{Proceedings of the IEEE/CVF Conference on Computer Vision
  and Pattern Recognition}, pages 14050--14059, 2021{\natexlab{a}}.

\bibitem[Radford et~al.(2021)Radford, Kim, Hallacy, Ramesh, Goh, Agarwal,
  Sastry, Askell, Mishkin, Clark, et~al.]{radford2021learning}
Alec Radford, Jong~Wook Kim, Chris Hallacy, Aditya Ramesh, Gabriel Goh,
  Sandhini Agarwal, Girish Sastry, Amanda Askell, Pamela Mishkin, Jack Clark,
  et~al.
\newblock Learning transferable visual models from natural language
  supervision.
\newblock \emph{arXiv preprint arXiv:2103.00020}, 2021.

\bibitem[Jia et~al.(2021)Jia, Yang, Xia, Chen, Parekh, Pham, Le, Sung, Li, and
  Duerig]{jia2021scaling}
Chao Jia, Yinfei Yang, Ye~Xia, Yi-Ting Chen, Zarana Parekh, Hieu Pham, Quoc~V
  Le, Yunhsuan Sung, Zhen Li, and Tom Duerig.
\newblock Scaling up visual and vision-language representation learning with
  noisy text supervision.
\newblock \emph{arXiv preprint arXiv:2102.05918}, 2021.

\bibitem[Oord et~al.(2018)Oord, Li, and Vinyals]{oord2018representation}
Aaron van~den Oord, Yazhe Li, and Oriol Vinyals.
\newblock Representation learning with contrastive predictive coding.
\newblock \emph{arXiv preprint arXiv:1807.03748}, 2018.

\bibitem[Li et~al.(2020)Li, Yin, Li, Zhang, Hu, Zhang, Wang, Hu, Dong, Wei,
  et~al.]{li2020oscar}
Xiujun Li, Xi~Yin, Chunyuan Li, Pengchuan Zhang, Xiaowei Hu, Lei Zhang, Lijuan
  Wang, Houdong Hu, Li~Dong, Furu Wei, et~al.
\newblock Oscar: Object-semantics aligned pre-training for vision-language
  tasks.
\newblock In \emph{European Conference on Computer Vision}, pages 121--137.
  Springer, 2020.

\bibitem[Zhang et~al.(2021)Zhang, Li, Hu, Yang, Zhang, Wang, Choi, and
  Gao]{zhang2021vinvl}
Pengchuan Zhang, Xiujun Li, Xiaowei Hu, Jianwei Yang, Lei Zhang, Lijuan Wang,
  Yejin Choi, and Jianfeng Gao.
\newblock Vinvl: Revisiting visual representations in vision-language models.
\newblock In \emph{Proceedings of the IEEE/CVF Conference on Computer Vision
  and Pattern Recognition}, pages 5579--5588, 2021.

\bibitem[Wang et~al.(2021{\natexlab{b}})Wang, Yu, Yu, Dai, Tsvetkov, and
  Cao]{wang2021simvlm}
Zirui Wang, Jiahui Yu, Adams~Wei Yu, Zihang Dai, Yulia Tsvetkov, and Yuan Cao.
\newblock Simvlm: Simple visual language model pretraining with weak
  supervision.
\newblock \emph{arXiv preprint arXiv:2108.10904}, 2021{\natexlab{b}}.

\bibitem[Hu et~al.(2021)Hu, Gan, Wang, Yang, Liu, Lu, and Wang]{hu2021scaling}
Xiaowei Hu, Zhe Gan, Jianfeng Wang, Zhengyuan Yang, Zicheng Liu, Yumao Lu, and
  Lijuan Wang.
\newblock Scaling up vision-language pre-training for image captioning.
\newblock \emph{arXiv preprint arXiv:2111.12233}, 2021.

\bibitem[Pennington et~al.(2014)Pennington, Socher, and
  Manning]{pennington2014glove}
Jeffrey Pennington, Richard Socher, and Christopher~D Manning.
\newblock Glove: Global vectors for word representation.
\newblock In \emph{Proceedings of the 2014 conference on empirical methods in
  natural language processing (EMNLP)}, pages 1532--1543, 2014.

\bibitem[Williams(1992)]{williams1992simple}
Ronald~J Williams.
\newblock Simple statistical gradient-following algorithms for connectionist
  reinforcement learning.
\newblock \emph{Machine learning}, 8\penalty0 (3):\penalty0 229--256, 1992.

\bibitem[Liu et~al.(2017)Liu, Zhu, Ye, Guadarrama, and Murphy]{liu2017improved}
Siqi Liu, Zhenhai Zhu, Ning Ye, Sergio Guadarrama, and Kevin Murphy.
\newblock Improved image captioning via policy gradient optimization of spider.
\newblock In \emph{Proceedings of the IEEE international conference on computer
  vision}, pages 873--881, 2017.

\bibitem[Devlin et~al.(2018)Devlin, Chang, Lee, and Toutanova]{devlin2018bert}
Jacob Devlin, Ming-Wei Chang, Kenton Lee, and Kristina Toutanova.
\newblock Bert: Pre-training of deep bidirectional transformers for language
  understanding.
\newblock \emph{arXiv preprint arXiv:1810.04805}, 2018.

\bibitem[Madhyastha et~al.(2019)Madhyastha, Wang, and
  Specia]{madhyastha2019vifidel}
Pranava Madhyastha, Josiah Wang, and Lucia Specia.
\newblock Vifidel: Evaluating the visual fidelity of image descriptions.
\newblock \emph{arXiv preprint arXiv:1907.09340}, 2019.

\bibitem[Kuznetsova et~al.(2020)Kuznetsova, Rom, Alldrin, Uijlings, Krasin,
  Pont-Tuset, Kamali, Popov, Malloci, Kolesnikov, et~al.]{kuznetsova2020open}
Alina Kuznetsova, Hassan Rom, Neil Alldrin, Jasper Uijlings, Ivan Krasin, Jordi
  Pont-Tuset, Shahab Kamali, Stefan Popov, Matteo Malloci, Alexander
  Kolesnikov, et~al.
\newblock The open images dataset v4.
\newblock \emph{International Journal of Computer Vision}, 128\penalty0
  (7):\penalty0 1956--1981, 2020.

\bibitem[Sharma et~al.(2018)Sharma, Ding, Goodman, and
  Soricut]{sharma2018conceptual}
Piyush Sharma, Nan Ding, Sebastian Goodman, and Radu Soricut.
\newblock Conceptual captions: A cleaned, hypernymed, image alt-text dataset
  for automatic image captioning.
\newblock In \emph{Proceedings of the 56th Annual Meeting of the Association
  for Computational Linguistics (Volume 1: Long Papers)}, pages 2556--2565,
  2018.

\bibitem[Papineni et~al.(2002)Papineni, Roukos, Ward, and
  Zhu]{papineni2002bleu}
Kishore Papineni, Salim Roukos, Todd Ward, and Wei-Jing Zhu.
\newblock Bleu: a method for automatic evaluation of machine translation.
\newblock In \emph{Proceedings of the 40th annual meeting of the Association
  for Computational Linguistics}, pages 311--318, 2002.

\bibitem[Lin(2004)]{lin2004rouge}
Chin-Yew Lin.
\newblock Rouge: A package for automatic evaluation of summaries.
\newblock In \emph{Text summarization branches out}, pages 74--81, 2004.

\bibitem[Banerjee and Lavie(2005)]{banerjee2005meteor}
Satanjeev Banerjee and Alon Lavie.
\newblock Meteor: An automatic metric for mt evaluation with improved
  correlation with human judgments.
\newblock In \emph{Proceedings of the acl workshop on intrinsic and extrinsic
  evaluation measures for machine translation and/or summarization}, pages
  65--72, 2005.

\bibitem[Anderson et~al.(2016{\natexlab{b}})Anderson, Fernando, Johnson, and
  Gould]{anderson2016spice}
Peter Anderson, Basura Fernando, Mark Johnson, and Stephen Gould.
\newblock Spice: Semantic propositional image caption evaluation.
\newblock In \emph{European conference on computer vision}, pages 382--398.
  Springer, 2016{\natexlab{b}}.

\bibitem[Changpinyo et~al.(2021)Changpinyo, Sharma, Ding, and
  Soricut]{changpinyo2021conceptual}
Soravit Changpinyo, Piyush Sharma, Nan Ding, and Radu Soricut.
\newblock Conceptual 12m: Pushing web-scale image-text pre-training to
  recognize long-tail visual concepts.
\newblock In \emph{Proceedings of the IEEE/CVF Conference on Computer Vision
  and Pattern Recognition}, pages 3558--3568, 2021.

\bibitem[Song et~al.(2020)Song, Liu, and Wang]{song2020revisiting}
Guanglu Song, Yu~Liu, and Xiaogang Wang.
\newblock Revisiting the sibling head in object detector.
\newblock In \emph{Proceedings of the IEEE/CVF Conference on Computer Vision
  and Pattern Recognition}, pages 11563--11572, 2020.

\bibitem[Lin et~al.(2014)Lin, Maire, Belongie, Hays, Perona, Ramanan,
  Doll{\'a}r, and Zitnick]{lin2014microsoft}
Tsung-Yi Lin, Michael Maire, Serge Belongie, James Hays, Pietro Perona, Deva
  Ramanan, Piotr Doll{\'a}r, and C~Lawrence Zitnick.
\newblock Microsoft coco: Common objects in context.
\newblock In \emph{European conference on computer vision}, pages 740--755.
  Springer, 2014.

\end{thebibliography}
